%% file: main.tex
\documentclass[10pt,twocolumn,letterpaper]{article}

\usepackage[pagenumbers]{wacv} % To force page numbers, e.g. for an arXiv version


\definecolor{wacvblue}{rgb}{0.21,0.49,0.74}
\usepackage[pagebackref,breaklinks,colorlinks,allcolors=wacvblue]{hyperref}
\usepackage{multirow}

\def\wacvPaperID{1105} % *** Enter the WACV Paper ID here
\def\confName{WACV}
\def\confYear{2027}

\title{SCLARO: A Dataset for Grounded Scenario-Level Scene Understanding and  ScenarioCLIP for Benchmarking}
\author{
  Advik Sinha \quad Saurabh Atreya \quad Aashutosh A V \quad Sk Aziz Ali \quad Abhijit Das \\
  Machine Intelligence Group, Department of CS\&IS, \\
  Birla Institute of Technology and Science, Pilani -- Hyderabad Campus \\
  {\tt\small abhijit.das@hyderabad.bits-pilani.ac.in}
}

\begin{document}
\maketitle
\input{sec/0_abstract}
\input{sec/1_intro}
\input{sec/2_formatting}
{
    \small
    \bibliographystyle{ieeenat_fullname}
    \bibliography{main}
}

% ─── Supplementary Material ───────────────────────────
\clearpage
\setcounter{section}{0}
\setcounter{figure}{0}
\setcounter{table}{0}
\renewcommand{\thesection}{\Alph{section}}
\renewcommand{\thefigure}{A\arabic{figure}}
\renewcommand{\thetable}{A\arabic{table}}

\twocolumn[
  {\centering\Large\textbf{Supplementary Material}\par\vspace{1em}}
]
\vspace{1em}

\appendix
\input{sec/A_impl_details}
\input{sec/B_data_curation}
\input{sec/C_experiment_details}

{
    \small
    \bibliographystyle{ieeenat_fullname}
    \bibliography{main}
}
\end{document}

%% file: sec/0_abstract.tex
\begin{abstract}
In the paradigm of computer vision-based precise real-world scene understanding, joint reasoning in terms of contextual understanding about the objects present in a scene, their inter-object relations, and the action being performed is an essential prerequisite. However, prior works have not addressed all three jointly, and no large-scale dataset provides grounded annotations at all three levels across diverse visual scenarios.
Hence, this work introduces
the SCLARO (\textbf{S}cene-\textbf{C}ontextual \textbf{L}ocalisation of \textbf{A}ctions, \textbf{R}elations \& \textbf{O}bjects) dataset, consisting of 615,805 images spanning indoor, outdoor, and driving scenarios, annotated with global action captions, object bounding boxes, and relation triplets that supply structured scene context beyond a free-text caption. To benchmark the dataset, we propose ScenarioCLIP, a tri-level reference model that jointly encodes global scene context, objects, and inter-object relations using
disentangled encoders and EMA-based knowledge distillation. We benchmark across a comprehensive suite of tasks on the SCLARO Dataset, namely zero-shot retrieval, linear probe, object detection, predicate classification, scene-graph classification, and out-of-domain generalisation. ScenarioCLIP's disentangled encoders improve over the previous works, such as PyramidCLIP's shared encoder, most notably at the object and relation levels and on out-of-domain generalisation. Code for the data generation pipeline and ScenarioCLIP is available at https://github.com/scenario-clip/SCLARO-ScenarioCLIP.
%The implementation for the dataset generation and benchmarking is available at https://github.com/scenario-clip/SCLARO-ScenarioCLIP, and the dataset is available on HuggingFace at https://huggingface.co/datasets/SCLARO/SCLARO.
\end{abstract}

%% file: sec/1_intro.tex
\begin{figure}[t]
    \centering
    \begin{subfigure}[b]{\columnwidth}
        \centering
        \includegraphics[width=\columnwidth]{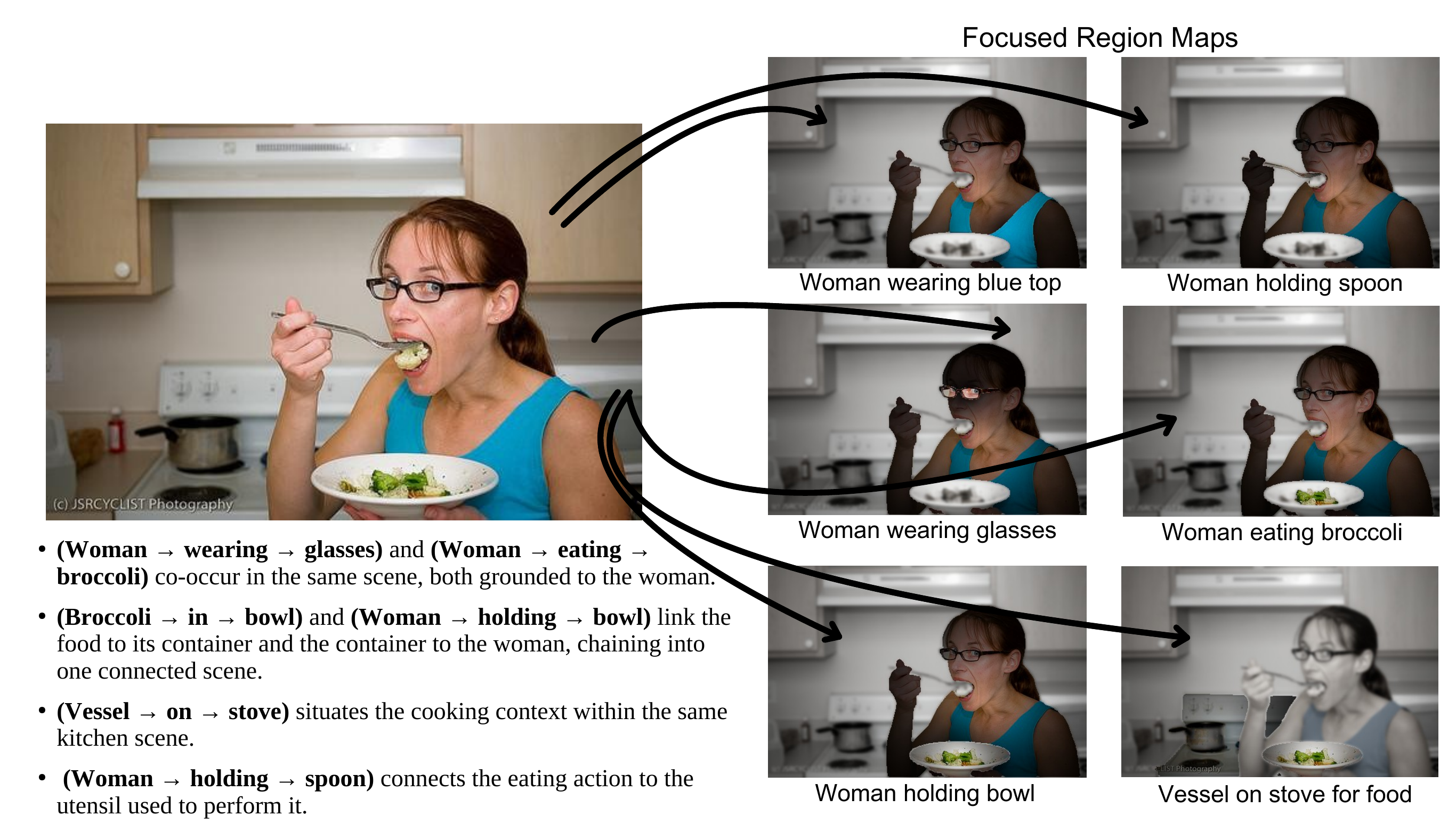}
        \caption{Relation-focused regions.}
        \label{fig:focused_regions}
    \end{subfigure}
    \\[1ex]
    \begin{subfigure}[b]{\columnwidth}
        \centering
        \includegraphics[width=\columnwidth]{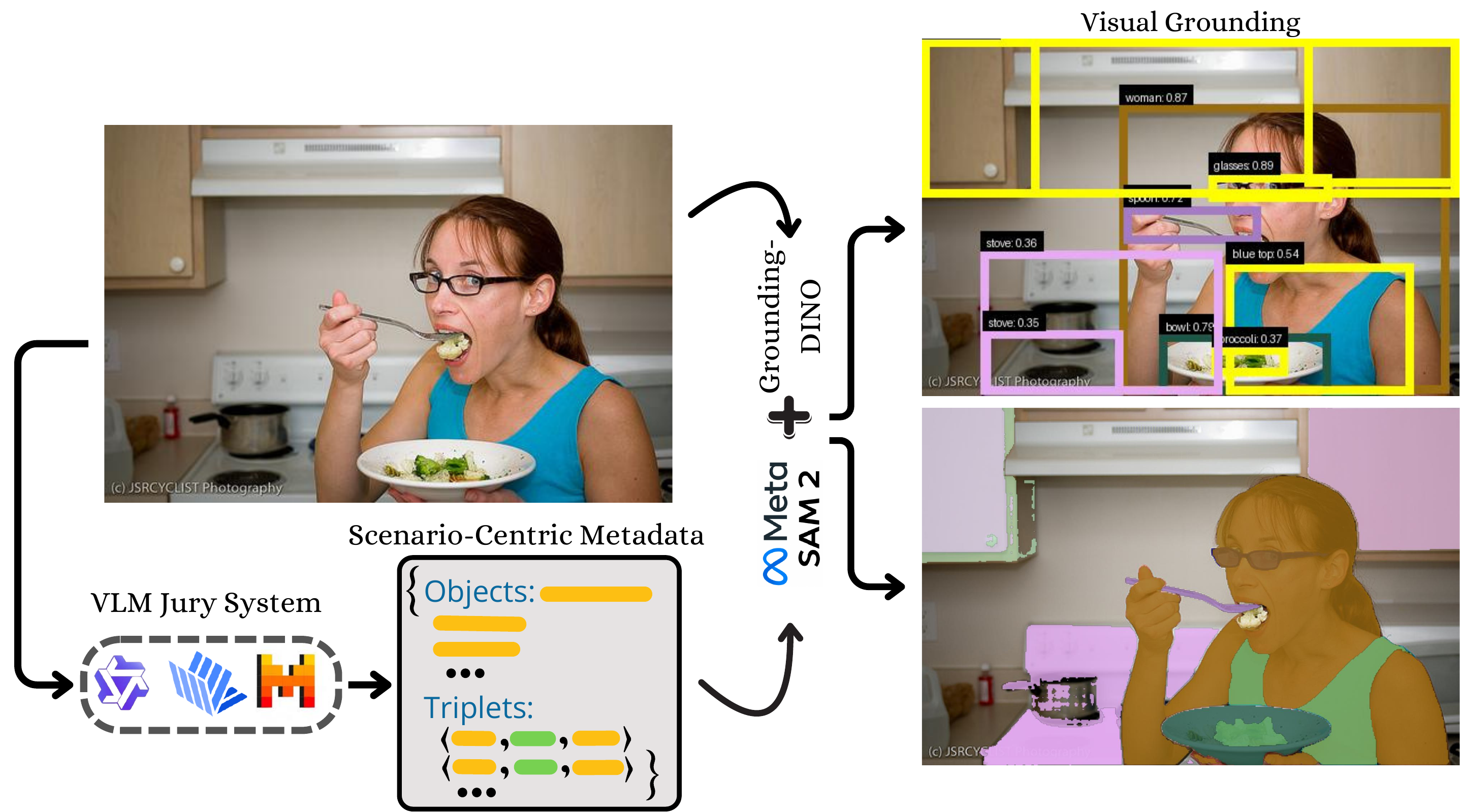}
        \caption{Scenario-centric annotation pipeline.}
        \label{fig:vlm_pipeline}
    \end{subfigure}
    \caption{\textbf{Focused regions and scenario-centric grounding.}
    (a) Our pipeline generates relation-focused regions for each triplet
    (e.g., \textit{woman wearing glasses}, \textit{vessel on stove for food}).
    (b) A VLM produces scenario-centric metadata,
    which GroundingDINO \cite{liu2023grounding} and SAM \cite{kirillov2023segment} then ground with bounding boxes and segmentation
    masks that we convert into the focused regions used by ScenarioCLIP.}
    \label{fig:focused_and_pipeline}
\end{figure}

\section{Introduction}
Real-world scene images exhibit rich compositional structure involving multiple objects, actions, and inter-object relations, building a complex semantics out of their context and location. However, existing large-scale vision-language pretraining datasets provide only global image-text pairs without grounded relational structure, while existing scene graph datasets~\cite{krishna2016visualgenomeconnectinglanguage,zhou2024openpsgopensetpanopticscene, hudson2019gqanewdatasetrealworld} lack the scale and annotation diversity required for VLP pretraining. This gap limits CLIP-type models' ability to reason about fine-grained scene contextual composition: embeddings of semantically similar pairs (e.g., \textit{person holding bag} vs. \textit{person holding umbrella}) collapse without grounded relational supervision at training time. Moreover, hierarchical CLIP-style models such as PyramidCLIP \cite{gao2022pyramidcliphierarchicalfeaturealignment} align global and object-level features but fall short of modelling inter-object relations, leaving open the question about how much contextual grounded relational supervision actually helps. 

To address this, we introduce the \emph{SCLARO} (\textbf{S}cene-\textbf{C}ontextual \textbf{L}ocalisation of \textbf{A}ctions, \textbf{R}elations \& \textbf{O}bjects) dataset, a 615,805-image corpus spanning indoor, outdoor, and driving scenarios. Each image is annotated with a global action caption, object bounding boxes, and relation-focused regions: grounded crops highlighting the visual evidence of each $(object_1, relation, object_2)$ triplet. These annotations are generated via a three-stage pipeline combining Ovis-Gemma~\cite{lu2024ovis} for structured scene annotation, GroundingDINO~\cite{liu2023grounding} for object localisation, and SAM~\cite{kirillov2023segment} with RBF-based weighting for focused region construction, across five publicly available sources: CC3M~\cite{sharma-etal-2018-conceptual}, OpenPSG~\cite{zhou2024openpsgopensetpanopticscene}, Panoptic Video Scene Graph~\cite{yang2023panopticvideoscenegraph}, Multi-Moments in Time~\cite{monfort2021multimomentstimelearninginterpreting}, and KITTI2015~\cite{geiger2012cvpr}.

To benchmark the SCLARO dataset and establish baseline performance for future work, we propose ScenarioCLIP, a hierarchical tri-level vision-language model trained on our dataset. ScenarioCLIP employs disentangled global, object, and relation encoders aligned via peer-level contrastive losses, with EMA-based knowledge distillation enforcing semantic consistency across hierarchical levels. As shown in Figure~\ref{fig:focused_and_pipeline}, the relation triplets accompanying each scene
supply explicit context as structured links between specific
objects, in a form usable
directly for relation-level supervision. %As illustrated in Figure~\ref{fig:focused_and_pipeline}, the SCLARO dataset decomposes each scene into relation-focused regions corresponding to grounded relation triplets
ScenarioCLIP leverages these pre-computed focused regions to enable explicit alignment between localised visual content and relation-specific text, a form of supervision absent in prior CLIP-type models, including PyramidCLIP~\cite{gao2022pyramidcliphierarchicalfeaturealignment}, which aligns global and local features but does not model inter-object relations. 

Trained and evaluated on the SCLARO dataset, ScenarioCLIP improves over the existing literature, such as PyramidCLIP \cite{gao2022pyramidcliphierarchicalfeaturealignment} across zero-shot retrieval, linear probe, object detection, predicate classification, and scene graph classification. External evaluation on MS-COCO \cite{lin2015microsoftcococommonobjects} and Visual Genome \cite{krishna2016visualgenomeconnectinglanguage} confirms that domain-specialised pretraining on the SCLARO dataset retains competitive general-purpose representations.

In summary, the contributions of this paper are:
\begin{itemize}
\item We introduce the \textit{SCLARO Dataset}, a 615,805-image corpus
    with grounded action captions, object bounding boxes, and
    relation-focused regions across five diverse visual domains, providing
    a new resource for scenario-level scene understanding.
    \item We propose \textit{ScenarioCLIP}, a tri-level vision-language
    model that jointly encodes global scene context, objects, and
    inter-object relations, and use it to test whether the SCLARO dataset's
    grounded, relation-level supervision improves representation quality
    over PyramidCLIP, the strongest prior model that aligns global and
    object-level features without modelling inter-object relations.
    \item We provide an evaluation protocol covering five tasks, namely,
    zero-shot retrieval, linear probe, object detection, predicate
    classification, and scene graph classification, evaluated across
    multiple models so that results reflect properties of the dataset
    itself, establishing the SCLARO dataset as a benchmark for future work
    on scenario-level scene understanding.
\end{itemize}

%\begin{figure}[t]
%    \centering
%    \begin{subfigure}[b]{0.45\columnwidth}
%        \centering
%        \includegraphics[width=\textwidth]{fig/single_relation.png}
%        \caption{Single relation}
%        \label{fig:single_relation}
%    \end{subfigure}
%    \hfill
%    \begin{subfigure}[b]{0.45\columnwidth}
%        \centering
%        \includegraphics[width=\textwidth]{fig/multi_relation.png}
%        \caption{Multiple relations}
%        \label{fig:multiple_relations}
%    \end{subfigure}
%    \caption{ScenarioCLIP can not only detect actions and objects but also localize the relations between objects in both single-relation (a) and multi-relation (b) scenes.}
%    \label{fig:relations}
%\end{figure}

\section{Previous Works} \paragraph{Vision-Language Pretraining.} Vision-Language Pretraining (VLP) models align visual and textual modalities to provide strong, transferable backbones for downstream tasks~\cite{schuhmann2021laion400mopendatasetclipfiltered,schuhmann2022laion5bopenlargescaledataset,zhou2024aligningmodalitiesvisionlarge,wang2024enhancingvisuallanguagemodalityalignment}. Pretraining has followed supervised~\cite{deng2009imagenet,girshick2015fastrcnn} and self-supervised~\cite{chen2020simpleframeworkcontrastivelearning,gui2024surveyselfsupervisedlearningalgorithms,he2020momentumcontrastunsupervisedvisual} regimes, with recent work using large VLMs to clean web-scale corpora for higher-quality CLIP-style training~\cite{wei2025hqclipleveraginglargevisionlanguage}. Dual-stream models such as CLIP~\cite{radford2021learningtransferablevisualmodels} dominate VLP by decoupling image and text encoders~\cite{jia2021scalingvisualvisionlanguagerepresentation,li2019visualbertsimpleperformantbaseline,li2019unicodervluniversalencodervision,lu2019vilbertpretrainingtaskagnosticvisiolinguistic}, enabling zero-shot transfer across classification~\cite{krizhevsky2009learning,sammani2024interpretinganalyzingclipszeroshot}, open-vocabulary segmentation~\cite{shao2024explorepotentialcliptrainingfree,zhu2024clipvisadaptingclipopenvocabulary,peng2025understanding}, action recognition~\cite{miech2020rareactvideodatasetunusual,wang2023seeingflowingadaptingclip}, and out-of-distribution detection~\cite{adaloglou2023adaptingcontrastivelanguageimagepretrained,wang2023clipnzeroshotooddetection}. CLIP has since been extended along many axes such as task-specific objectives and high-capacity designs~\cite{yao2021filipfinegrainedinteractivelanguageimage,yu2022cocacontrastivecaptionersimagetext,zhang2024visionlanguagemodelsvisiontasks}, lightweight adaptation and prompting~\cite{gao2021clipadapterbettervisionlanguagemodels,zhang2021tipadaptertrainingfreeclipadapterbetter,Zhou_2022,zhou2022conditionalpromptlearningvisionlanguage,chen2023clipguidedimageperceptiveprompt}, knowledge distillation for open-vocabulary detection and segmentation~\cite{ding2022decouplingzeroshotsemanticsegmentation,du2022learningpromptopenvocabularyobject,gu2022openvocabularyobjectdetectionvision}, modality-gap and continual-learning post-training~\cite{yamaguchi2025postpretrainingmodalityalignmentvisionlanguage,liu2025cclip}, video~\cite{wang2021actionclipnewparadigmvideo,wang2023videocomposercompositionalvideosynthesis}, generation~\cite{wang2022clipgenlanguagefreetrainingtexttoimage,li2023blipdiffusionpretrainedsubjectrepresentation}, captioning and segmentation~\cite{xu2022groupvitsemanticsegmentationemerges,lai2024veclipimprovingcliptraining}, driving~\cite{xu2024drivegpt4interpretableendtoendautonomous}, 
and multimodal instruction following~\cite{liu2023visualinstructiontuning,mckinzie2024mm1methodsanalysis}. \\
\textbf{Hierarchical and Relational CLIP.} Most relevant to our setting are models that move beyond global image-text alignment toward finer structure. PyramidCLIP~\cite{gao2022pyramidcliphierarchicalfeaturealignment} aligns global images with localised sub-regions across a vision-language pyramid, and FocusCLIP~\cite{khan2024humanposedescriptionssubjectfocused} and Eyes Wide Shut~\cite{tong2024eyeswideshutexploring} sharpen attention to salient regions. These, however are restricted to the object/region level and do not model inter-object relations. A separate line targets relational and compositional structure: TripletCLIP~\cite{patel2024tripletclipimprovingcompositionalreasoning} improves compositional reasoning with synthetic hard negatives, Structure-CLIP~\cite{huang2023structureclipscenegraphknowledge} injects scene-graph knowledge but relies on external knowledge graphs rather than grounded regions, and SGVL~\cite{herzig2023incorporatingstructuredrepresentationspretrained} adds scene-graph supervision but operates on existing Visual Genome annotations rather than a purpose-built grounded corpus. Recent compositional VLP~\cite{li2025enhancingvisionlanguagecompositionalunderstanding} further motivates large-scale grounded relation data. \\
%In contrast to all of these, our dataset aligns relation-specific text with spatially grounded relation regions at pretraining scale, and our benchmark evaluates representations jointly at the action, object, and relation levels -- supervision and evaluation absent from prior CLIP-style work. 
\textbf{Scene Graph Datasets.} Several datasets target structured scene understanding. Visual Genome~\cite{krishna2016visualgenomeconnectinglanguage} provides dense object, attribute, and relation annotations over 108K images, but is crowdsourced and lacks action captions or focused relation regions. GQA~\cite{hudson2019gqanewdatasetrealworld} repurposes Visual Genome for compositional question answering but does not support VLP pretraining. OpenPSG~\cite{zhou2024openpsgopensetpanopticscene} introduces panoptic scene graphs but is limited in scale, and video-based datasets such as Action Genome~\cite{Ji_2020_CVPR} and Panoptic Video Scene Graph~\cite{yang2023panopticvideoscenegraph} provide spatio-temporal relations restricted to specific video corpora. In contrast, the SCLARO dataset provides grounded relation annotations at the VLP scale across diverse visual domains, pairing relation-focused regions with action captions designed specifically for vision-language pretraining. Prior CLIP-style models lack grounded relation supervision, and prior scene graph datasets lack pretraining-scale action, object, and relation annotations together. To mitigate these gaps, the SCLARO dataset addresses both gaps, aligning relation-specific text with spatially grounded relation regions at
pretraining scale, while our benchmark jointly evaluates representations
at the action, object, and relation levels.

\begin{figure*}[ht!]
  \centering
  \includegraphics[width=0.7\textwidth]{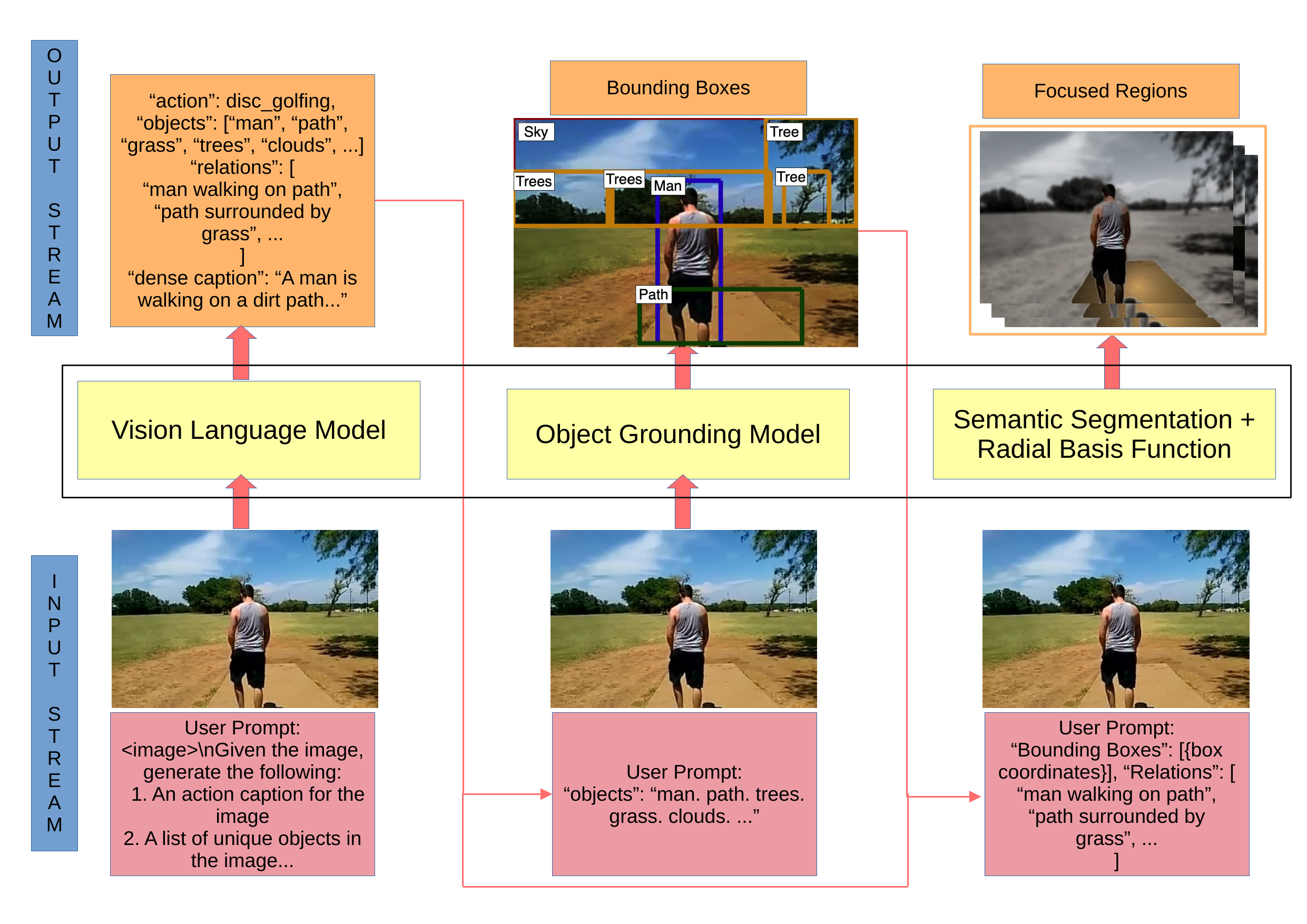}
  \hfill
  \caption{\textbf{Data generation pipeline for the SCLARO Dataset.} Given a raw image, a vision-language model produces a global action caption, object list, and relation triplets (left). An object grounding model (GroundingDINO~\cite{liu2023grounding}) then predicts bounding boxes for the mentioned objects (middle). Finally, a segmentation model (SAM~\cite{kirillov2023segment}) with RBF-based weighting constructs relation-focused regions that highlight the spatial context of each $(object_1, relation, object_2)$ triplet (right).}
  \label{fig:new_arch}
\end{figure*}

\section{SCLARO Dataset}\label{AGD}

The SCLARO dataset is a large-scale image dataset comprising 615,805 images spanning indoor, outdoor, and driving scenarios. Each data point is structured with multiple layers of descriptive and relational information: a global action caption, a list of object names with bounding boxes, and a set of relation triplets, each linked to a spatially grounded focused region. Together these components provide a comprehensive, multifaceted view of actions, objects, and relationships within visual scenes.

\textbf{Stage 1 - Structured Scene Annotation.} Each image is annotated with a main action label and a concise text caption describing the primary action occurring within the scene. Each image also includes a list of objects identifying all visible entities, where a single object type may appear multiple times, denoting distinct instances. Additionally, each image includes a set of relation triplets structured as $(object_1, relation, object_2)$, where $relation$ indicates a specific interaction or association between $object_1$ and $object_2$. Each image contains at least three such triplets, offering rich relational data that enables fine-grained understanding of object interactions within the scene.

\textbf{Stage 2 - Object Localisation.} For each identified object, the dataset provides precise bounding boxes $(x_{\text{min}}, y_{\text{min}}, x_{\text{max}}, y_{\text{max}})$ that define the object's location within the image.

\textbf{Stage 3 - Focused Region Construction.} Each relation triplet is supplemented with a focused region, a specific area of the image that highlights the spatial context of the respective relation. These regions emphasise the segments of the image where the interaction between $object_1$ and $object_2$ is most pronounced, enabling explicit visual grounding of relational evidence.

\subsection{Models used for Stage-wise Generation}
Stage-1 uses an open-weight VLM (Ovis-Gemma 9B~\cite{gemmateam2024gemma2improvingopen,lu2024ovis}) to jointly generate a global action caption, dense caption, object list, and relation triplets for each image within a single prompt (\cref{fig:new_arch}). The relation triplets supply scene context in a structured, queryable form, with explicit links between objects, that can be used directly for grounded relation-level supervision.

Stage-2 applies GroundingDINO~\cite{liu2023grounding}, prompted with the object list, to predict bounding boxes for all mentioned objects.
Stage-3 uses SAM~\cite{kirillov2023segment} to obtain object masks and constructs relation-focused regions by spatially weighting and blending the masks with the original image using RBF-based weighting.

Further implementation details (prompt design and the exact RBF/Gaussian weighting used to form focused regions) are provided in Sec.~A of the supplementary material.

% \begin{figure}[h]
% \centering
% \includegraphics[width=0.7\columnwidth]{fig/size_pie.png}
% \caption{Domain breakdown of the SCLARO Dataset across constituent sources.}
% \label{fig:dataset_breakdown}
% \end{figure}

\subsection{Constituent Datasets and Curation}\label{sec:datacur}
The SCLARO dataset is constructed from five publicly available sources:
CC3M~\cite{sharma-etal-2018-conceptual},
OpenPSG~\cite{zhou2024openpsgopensetpanopticscene},
Panoptic Video Scene Graph (PVSG)~\cite{yang2023panopticvideoscenegraph},
Multi-Moments in Time~\cite{monfort2021multimomentstimelearninginterpreting},
and KITTI2015~\cite{geiger2012cvpr}.
Wherever available, ground-truth labels are reused (e.g., action classes in Multi-Moments in Time, and relations in OpenPSG and PVSG).

To obtain a clean label space, we normalise and consolidate the raw action, object, and relation labels using a combination of automatic filtering and clustering, as well as manual curation, yielding condensed sets of action, object, and relation classes used in all our experiments. During pipeline construction, each annotation stage was subject to manual review. Annotations identified as incorrect (e.g., hallucinated objects, semantically invalid relations) were excluded from the final dataset. Action and object annotations required exclusion only rarely, while relations were excluded somewhat more frequently, reflecting the comparatively greater difficulty of grounding open-vocabulary relations. Additional details of the curation pipeline are provided in Sec.~B of the supplementary material. 

The proposed SCLARO dataset contains \textbf{615{,}805} images, \textbf{740} action classes, \textbf{4{,}812} object classes, and \textbf{225{,}609} unique $(object_1, predicate, object_2)$ relation triplets. 
% The number of unique $predicates$ in the dataset is \textbf{8{,}551}. // better framing

\begin{table}[h]
\centering
\small
\begin{tabular}{lrr}
\toprule
\textbf{Source Dataset} & \textbf{Images} & \textbf{\% of Total} \\
\midrule
CC3M & 264,838 & 43.0\% \\
Multi-Moments in Time & 226,066 & 36.7\% \\
OpenPSG &  62,685 & 10.2\% \\
Panoptic Video Scene Graph &  54,223 &  8.8\% \\
KITTI2015 &   7,993 &  1.3\% \\
\midrule
\textbf{Total}                                                     & \textbf{615,805} & \textbf{100\%} \\
\bottomrule
\end{tabular}
\caption{Domain breakdown of the SCLARO Dataset across constituent sources.}
\label{tab:dataset_breakdown}
\end{table}

\begin{figure*}[t]
    \centering
    \includegraphics[width=\textwidth]{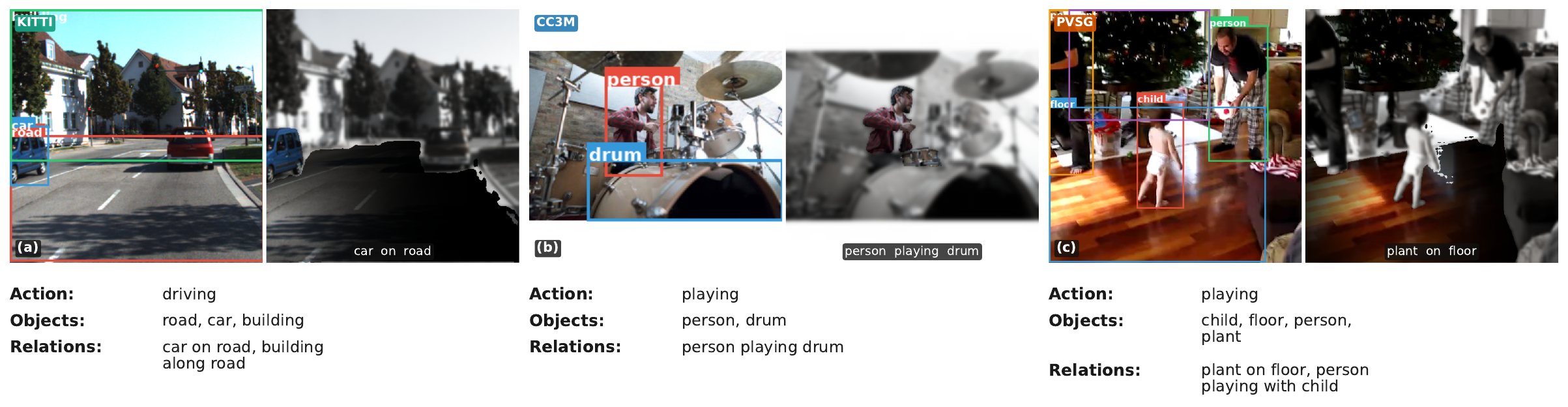}
    \caption{\textbf{Representative annotated examples from the SCLARO
    Dataset.} For each example, the left image shows the source image
    with object bounding boxes, and the right image shows a
    relation-focused region with the corresponding relation triplet
    overlaid, for one of the annotated relations in the image: (a)
    KITTI2015, car on road, (b) CC3M, person playing drum, (c) PVSG,
    plant on floor. The full action caption, object list, and relation
    triplets are shown below each panel.}
    \label{fig:main_qual}
\end{figure*}

\subsection{Dataset Analysis}

\Cref{tab:dataset_breakdown} shows the distribution of images across the five constituent source datasets. CC3M~\cite{sharma-etal-2018-conceptual} and Multi-Moments in Time~\cite{monfort2021multimomentstimelearninginterpreting} contribute the most images (79.7\%), providing broad action and scene diversity. OpenPSG~\cite{zhou2024openpsgopensetpanopticscene} and Panoptic Video Scene Graph~\cite{yang2023panopticvideoscenegraph} provide relation-rich scene graph supervision (19.0\%), while KITTI2015~\cite{geiger2012cvpr} contributes outdoor driving scenarios, a distinct visual domain (1.3\%). The relation vocabulary exhibits a long-tailed distribution, with the top 10\% of predicates accounting for 95.7\% of relation instances, and 41.3\% appearing only once. This reflects the diversity of real-world relations captured by our open-vocabulary pipeline, and is consistent with relation distributions in other large-scale scene graph datasets such as Visual Genome~\cite{krishna2016visualgenomeconnectinglanguage}.

\Cref{fig:main_qual} shows three representative annotated examples spanning multiple sources, illustrating the dataset's grounded annotations. Additional qualitative examples are provided in Sec. C of the supplementary material.

We do not redistribute source images, drawn from publicly available datasets and subject to their original licenses. Each annotation references its source dataset and image identifier, so users can obtain images directly from that source. We release the action, object, and relation annotations, along with the relation-focused region crops.

%\subsection{Annotation Quality}
%To assess annotation quality, we manually evaluated 500 
%randomly sampled images (0.08\% of the dataset). For each image, 
%the action caption was rated correct if it accurately described the 
%primary activity, object labels were rated correct if all listed 
%objects were visible in the image, and relation triplets were rated 
%correct if the stated interaction was semantically accurate. Action 
%captions achieved 85.3\% precision, object labels 90.7\%, and 
%relation triplets 86.6\%, confirming that the automated pipeline 
%produces reliable annotations across all three levels of scene 
%description (see the supplementary material (Appendix~C) for qualitative examples).

%% file: sec/2_formatting.tex
\section{ScenarioCLIP}

\begin{figure*}[ht!]
  \centering
  \includegraphics[width=1.0\textwidth, height=0.475\textheight, keepaspectratio]{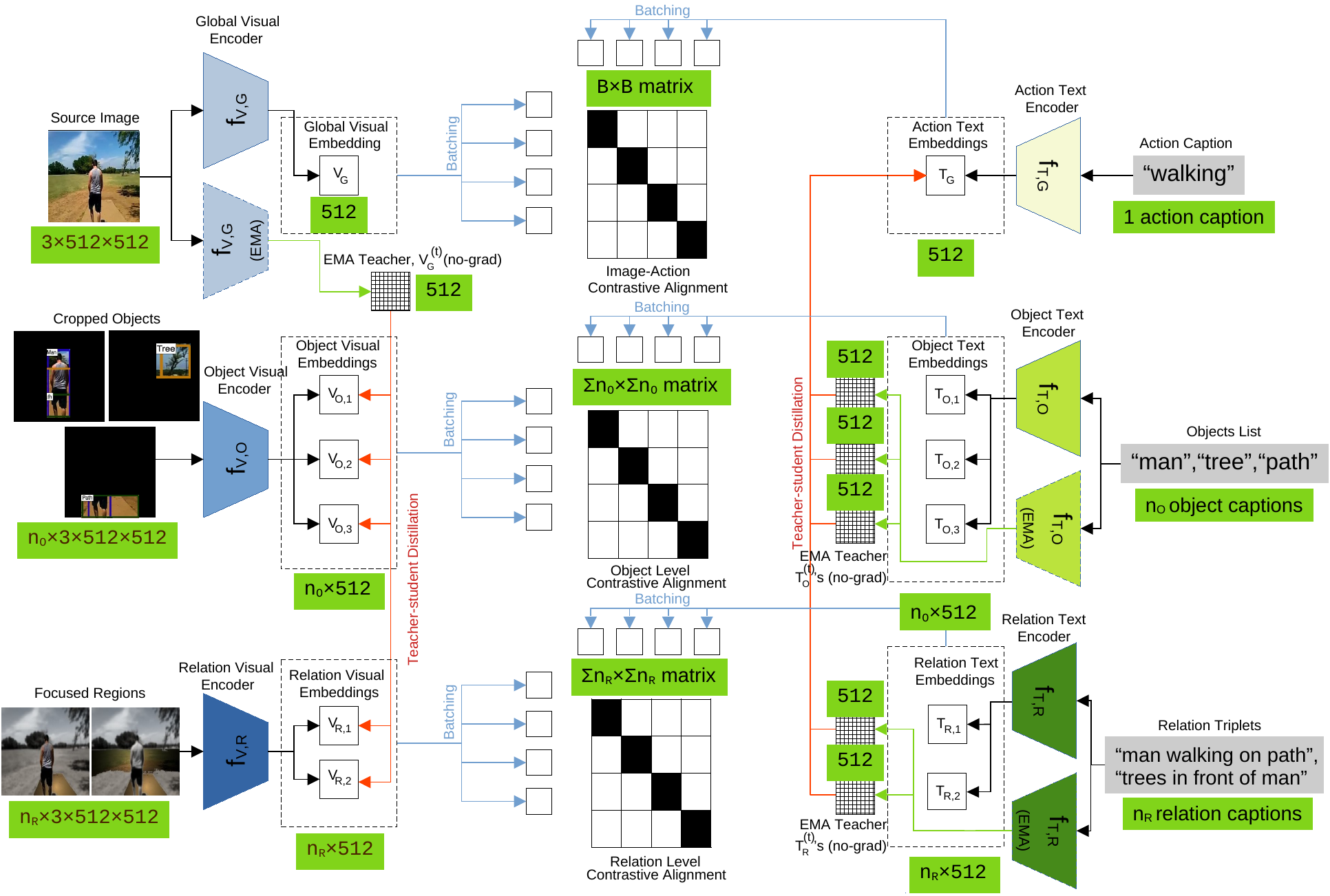}
  \caption{\textbf{Overview of ScenarioCLIP.} A global, object, and 
  relation encoder extracts visual features from the full image, object 
  crops, and focused regions, while corresponding text encoders embed 
  the action caption, object names, and relation triplets. Contrastive 
  losses align image-action, object-object, and relation-relation pairs, 
  and EMA teachers provide knowledge-distillation targets.}
  \label{fig:main_drawing}
\end{figure*}

To demonstrate SCLARO's utility and establish baseline
performance for future work, we propose ScenarioCLIP, a hierarchical
tri-level VLM trained and evaluated on the dataset introduced in
\cref{AGD}. ScenarioCLIP consists of three distinct visual encoders
$f_{V,G}$, $f_{V,O}$, $f_{V,R}$ and three distinct text encoders
$f_{T,G}$, $f_{T,O}$, $f_{T,R}$, extracting features at the
global, object, and relation levels respectively (\cref{fig:main_drawing}).
All encoders are initialised with pretrained CLIP weights
(\texttt{openai/clip-vit-base-patch32}~\cite{radford2021learningtransferablevisualmodels}),
providing a strong general-purpose prior that is specialised for scene-,
object-, and relation-level semantics during training on the SCLARO
dataset. The global encoder is trained on the full image, while the
object and relation encoders are trained on single-object crops and
relation-focused regions respectively, specialising them for
fine-grained recognition.

In contrast, PyramidCLIP \cite{gao2022pyramidcliphierarchicalfeaturealignment} routes the full image, object crops, and relation-focused regions through one shared visual encoder, and the text inputs through a shared text encoder. PyramidCLIP \cite{gao2022pyramidcliphierarchicalfeaturealignment} is trained with a contrastive
objective over this shared encoder's output. ScenarioCLIP, in contrast,
is trained with two objectives: a contrastive alignment loss
$\mathcal{L}_{CA}$ that aligns visual and text embeddings at each level,
and a knowledge distillation loss $\mathcal{L}_{KD}$, that acts as a regulariser across levels.

\subsection{Knowledge Distillation}\label{kd}
We employ knowledge distillation as a regulariser: at each step, an EMA
teacher provides stable targets that constrain the
geometry of the per-level embedding spaces, improving generalisation and
representation quality. The teacher is updated as a slow-moving average
of the student with decay $0.9995$ and a $2000$-step warm-up. For the visual modality, knowledge is transferred from the EMA global
visual teacher ${V_G}^{(t)}$ to each object-level embedding $V_{O,i}$
and relation-level embedding $V_{R,i}$, using Kullback-Leibler
divergence with frozen teacher targets detached from the computational
graph. On the text side, the direction is inverted: each EMA
object-level embedding ${T_{O,i}}^{(t)}$ and relation-level embedding
${T_{R,i}}^{(t)}$ serves as teacher for the global text embedding $T_G$.
We provide the precise distillation formula, including the softmax
temperature parameterisation, in Sec.~D.3 of the supplementary material. 

The distillation direction is asymmetric and intra-modal. On the visual
side, the global image carries the broadest contextual signal, so
$V_G^{(t)}$ regularises the localised object and relation embeddings,
which observes only partial scene context. On the text side, the direction
inverts: fine-grained descriptions (\textit{person wearing glasses},
\textit{vessel on stove}) specify content more precisely than the global
action caption, so the object- and relation-level text embeddings
regularise the global text embedding $T_G$.

% This asymmetric design is deliberate. On the visual side, the global image captures the broadest contextual signal, making ${V_G}^{(t)}$ a natural teacher for localised crops that observe only partial scene context. On the text side, fine-grained descriptions such as \textit{person wearing glasses} and \textit{vessel on stove} more precisely specify scene content than the global action caption alone, making ${T_{O,i}}^{(t)}$ and ${T_{R,j}}^{(t)}$ richer teachers for $T_G$. This asymmetry reflects the complementary nature of visual and linguistic hierarchies in scene understanding.

\begin{table*}[!t]
\tiny
  \centering
  \small
  \setlength{\tabcolsep}{4pt}
  
  \begin{tabular}{lccccccccc}
    \toprule
    & \multicolumn{3}{c}{\textbf{Actions}} & \multicolumn{3}{c}{\textbf{Objects}} & \multicolumn{3}{c}{\textbf{Relations}} \\
    \cmidrule(lr){2-4}\cmidrule(lr){5-7}\cmidrule(lr){8-10}
    \textbf{Model} & Top-1 & Top-5 & Top-10 & Top-1 & Top-5 & Top-10 & Top-1 & Top-5 & Top-10 \\
    \midrule
    \multicolumn{10}{l}{\emph{A. Ours (in-domain pretraining on SCLARO dataset)}} \\
    PyramidCLIP~\cite{gao2022pyramidcliphierarchicalfeaturealignment} (NeurIPS, 2022)  & 54.95 & 73.36 & 78.84 & 37.67 & 59.58 & 65.92 & 16.43 & 34.27 & 44.20 \\
    ScenarioCLIP w/o KD & 57.52 & \textbf{75.24} & \textbf{80.34} & 51.26 & 64.37 & \textbf{68.32} & 19.45 & 36.94 & 45.75 \\
    ScenarioCLIP & \textbf{57.63} & 75.02 & 79.99 & \textbf{51.86} & \textbf{64.43} & 68.30 & \textbf{19.56} & \textbf{37.03} & \textbf{45.86} \\
    \midrule
    \multicolumn{10}{l}{\emph{B. SoTA zero-shot baselines (released weights; prompt-only)}} \\
    CLIP~\cite{radford2021learningtransferablevisualmodels} (ICML, 2021)      & 14.87 & 33.43 & 42.45 &  3.77 & 12.35 & 18.39 &  0.66 &  2.67 &  4.72 \\
    SLIP~\cite{mu2021slipselfsupervisionmeetslanguageimage} (ECCV, 2022)      &  8.31 & 21.83 & 30.26 &  2.96 &  9.60 & 14.22 &  0.35 &  1.63 &  3.06 \\
    TripletCLIP~\cite{patel2024tripletclipimprovingcompositionalreasoning} (NeurIPS, 2024) &  5.48 & 14.91 & 21.67 &  0.53 &  2.21 &  3.84 &  0.03 &  0.14 &  0.27 \\
    FineCLIP~\cite{3737916.3738791} (NeurIPS, 2024) & 14.76 & 32.88 & 42.00 & 7.76 & 20.11 & 26.88 & 0.86 & 3.75 & 6.63 \\
    \bottomrule
  \end{tabular}
  \caption{Zero-shot (ZS) image$\rightarrow$text retrieval on the SCLARO Dataset.}
  \label{tab:zs_retrieval_full}
\end{table*}

\begin{table*}[t]
  \centering
  \small
  \setlength{\tabcolsep}{4pt}
  \begin{tabular}{lccccccccc}
    \toprule
    & \multicolumn{3}{c}{\textbf{Actions}} & \multicolumn{3}{c}{\textbf{Objects}} & \multicolumn{3}{c}{\textbf{Relations}} \\
    \cmidrule(lr){2-4}\cmidrule(lr){5-7}\cmidrule(lr){8-10}
    \textbf{Model} & Top-1 & Top-5 & Top-10 & Top-1 & Top-5 & Top-10 & Top-1 & Top-5 & Top-10 \\
    \midrule
    PyramidCLIP \cite{gao2022pyramidcliphierarchicalfeaturealignment}  & 75.36 & 90.54 & 93.52 & 79.74 & 91.79 & \textbf{94.37} & 40.78 & 59.73 & 66.09 \\
    ScenarioCLIP w/o KD  & \textbf{79.46} & \textbf{92.94} & \textbf{95.18} & 84.25 & \textbf{92.38} & 94.25 & \textbf{44.75} & 61.02 & 66.86 \\
    ScenarioCLIP  & 79.19 & 92.81 & 95.10 & \textbf{84.30} & 92.31 & 94.24 & 44.71 & \textbf{61.09} & \textbf{66.88} \\
    \bottomrule
  \end{tabular}
  \caption{Linear-probe results on the SCLARO Dataset.}
  \label{tab:lp_ablation}
\end{table*}

\subsection{Contrastive Alignment}

Cross-modal alignment is achieved by applying contrastive loss at each 
corresponding level: between $V_G$ and $T_G$, between $V_{O,i}$ and 
$T_{O,i}$, and between $V_{R,i}$ and $T_{R,i}$. We use the symmetric 
image-text contrastive objective of 
CLIP~\cite{radford2021learningtransferablevisualmodels} with a learnable 
temperature. The alignment matrix sizes at the global, object, and 
relation levels are $B \times B$, $\sum^{B} n_O \times \sum^{B} n_O$, 
and $\sum^{B} n_R \times \sum^{B} n_R$ respectively.

\subsection{Objective Function}

The total objective combines knowledge distillation and contrastive 
alignment:

\begin{equation}
    \mathcal{L}_{\text{total}} = \lambda_{KD} \cdot \mathcal{L}_{KD} 
    + \lambda_{CA} \cdot \mathcal{L}_{CA},
    \label{eq:objective}
\end{equation}

where $\lambda_{KD}$ and $\lambda_{CA}$ are weighting coefficients. The 
knowledge distillation loss $\mathcal{L}_{KD}$ aligns global embeddings 
with object and relation embeddings within each modality:

\begin{multline}
\mathcal{L}_{KD} = \sum_{i=1}^{n_O} D_{KL}({V_G}^{(t)} \| V_{O,i})
+ \sum_{j=1}^{n_R} D_{KL}({V_G}^{(t)} \| V_{R,j}) \\
+ \sum_{i=1}^{n_O} D_{KL}({T_{O,i}}^{(t)} \| T_G)
+ \sum_{j=1}^{n_R} D_{KL}({T_{R,j}}^{(t)} \| T_G)
\end{multline}

where $D_{KL}(\cdot \| \cdot)$ denotes KL divergence with frozen EMA 
teacher targets. The contrastive alignment loss $\mathcal{L}_{CA}$ 
encourages alignment between corresponding visual and textual embeddings 
across all three levels:

\begin{multline}
\mathcal{L}_{CA} = \text{Contrastive}(V_G, T_G)
+ \sum_{i=1}^{n_O} \text{Contrastive}(V_{O,i}, T_{O,i}) \\
+ \sum_{j=1}^{n_R} \text{Contrastive}(V_{R,j}, T_{R,j})
\end{multline}

% \begin{table}[h]
% \centering
% \small
% \caption{Predicate classification (PredCls) and scene graph classification (SGCls) results, reporting mean Recall@$K$ (mR@$K$) }
% % over the 1,320-predicate evaluation vocabulary ($\geq$5 test instances, 98.9\% coverage). |
% \begin{tabular}{l ccc ccc}
% \toprule
% & \multicolumn{3}{c}{PredCls} & \multicolumn{3}{c}{SGCls} \\
% \cmidrule(lr){2-4}\cmidrule(lr){5-7}
% \textbf{Model} & mR@1 & mR@5 & mR@10 & mR@1 & mR@5 & mR@10 \\
% \midrule
% PyramidCLIP~\cite{gao2022pyramidcliphierarchicalfeaturealignment} & 7.52 & 20.57 & 28.33 & 6.56 & 18.01 & 25.57 \\
% ScenarioCLIP w/o KD & 9.01 & \textbf{22.81} & \textbf{31.16} & 8.18 & 20.92 & 28.65 \\
% ScenarioCLIP & \textbf{9.05} & 22.67 & 30.60 & \textbf{8.38} & \textbf{21.23} & \textbf{28.73} \\
% \bottomrule
% \end{tabular}
% \label{tab:pred-sgcls}
% \end{table}

\begin{table}[h]
\centering
\setlength{\tabcolsep}{1.6pt}
\footnotesize
\begin{tabular}{l ccc ccc}
\toprule
& \multicolumn{3}{c}{PredCls} & \multicolumn{3}{c}{SGCls} \\
\cmidrule(lr){2-4}\cmidrule(lr){5-7}
\textbf{Model} & mR@1 & mR@5 & mR@10 & mR@1 & mR@5 & mR@10 \\
\midrule
PyramidCLIP~\cite{gao2022pyramidcliphierarchicalfeaturealignment} & 7.52 & 20.57 & 28.33 & 6.56 & 18.01 & 25.57 \\
ScenarioCLIP w/o KD & 9.01 & \textbf{22.81} & \textbf{31.16} & 8.18 & 20.92 & 28.65 \\
ScenarioCLIP & \textbf{9.05} & 22.67 & 30.60 & \textbf{8.38} & \textbf{21.23} & \textbf{28.73} \\
\bottomrule
\end{tabular}
\caption{Predicate classification (PredCls) and scene graph classification (SGCls) results, reporting mean Recall@$K$ (mR@$K$).}
\label{tab:pred-sgcls}
\end{table}

\section{Experiments}
\subsection{Implementation Details}
 
\noindent \textbf{Pre-training.} Models are pretrained on the
SCLARO dataset introduced in \cref{AGD}. We use an
AdamW~\cite{loshchilov2019decoupledweightdecayregularization} optimiser
with a linear warm-up over the first $10\%$ of steps followed by cosine
annealing. For ScenarioCLIP, the distillation temperature is optimised
at one-tenth the base learning rate for stability. All reported results
use $\lambda_{KD} = 1$ and $\lambda_{CA} = 1$ in
\cref{eq:objective}. Additional details are provided in Sec. D.1 of the
supplementary material.
 
\noindent \textbf{Downstream Tasks.} We benchmark the SCLARO dataset
across a comprehensive suite of tasks to assess its value as a resource
for improving scene understanding and representation: zero-shot
image$\rightarrow$text retrieval, linear probe, object detection,
predicate classification, and scene-graph classification on the SCLARO
dataset test set, as well as out-of-domain object classification on
MS-COCO~\cite{lin2015microsoftcococommonobjects} and Visual
Genome~\cite{krishna2016visualgenomeconnectinglanguage}, and
out-of-domain relation zero-shot image$\rightarrow$text retrieval on
Visual Genome~\cite{krishna2016visualgenomeconnectinglanguage}. Each
task is evaluated across three models, PyramidCLIP, ScenarioCLIP
w/o KD, and ScenarioCLIP, so that the results reflect properties of the
dataset itself rather than any single architecture's behaviour.
 
% [Table tab:pred-sgcls stays here, unchanged]
 
\subsection{Zero-Shot Retrieval}
Given an image as a query, we rank a fixed set of class strings using
cosine similarity between $\ell_2$-normalised image and text embeddings,
with no task-specific tuning. The candidate set at each level is the
full corresponding vocabulary introduced in \cref{sec:datacur}: 740
action classes, 4,812 object classes, and 225,609 unique
$(object_1, predicate, object_2)$ relation triplets. We report this task
across PyramidCLIP, ScenarioCLIP w/o KD, ScenarioCLIP, and several
publicly available zero-shot baselines
(CLIP~\cite{radford2021learningtransferablevisualmodels},
SLIP~\cite{mu2021slipselfsupervisionmeetslanguageimage},
TripletCLIP~\cite{patel2024tripletclipimprovingcompositionalreasoning},
and FineCLIP~\cite{3737916.3738791}, using averaged prompt templates)
(\cref{tab:zs_retrieval_full}). All three SCLARO dataset-trained models
substantially outperform the publicly available baselines.
 
ScenarioCLIP improves over
PyramidCLIP~\cite{gao2022pyramidcliphierarchicalfeaturealignment} at all
three levels, with the largest gains on \textit{Objects}
($14.19\%/4.85\%/2.38\%$ on Top-$1/5/10$) and \textit{Relations}
($3.13\%/2.76\%/1.66\%$), and a smaller gain on \textit{Actions}
($2.68\%/1.66\%/1.15\%$). Separate encoders for each level help
ScenarioCLIP learn better representations than PyramidCLIP's single
shared encoder.
Notably, TripletCLIP~\cite{patel2024tripletclipimprovingcompositionalreasoning}
underperforms CLIP~\cite{radford2021learningtransferablevisualmodels} on
SCLARO retrieval, indicating that hard-negative fine-tuning distorts
the embedding space in ways that do not transfer to scenario-level
retrieval on a novel domain.
Zero-shot retrieval difficulty also varies substantially across the
five constituent sources, from single-digit Relation Top-$1$
accuracy on CC3M~\cite{sharma-etal-2018-conceptual}, Multi-Moments in
Time~\cite{monfort2021multimomentstimelearninginterpreting}, and
OpenPSG~\cite{zhou2024openpsgopensetpanopticscene} to over $80\%$ on
PVSG~\cite{yang2023panopticvideoscenegraph}, with
KITTI2015~\cite{geiger2012cvpr} in between. The same relative ordering
holds for actions and objects, and critically, this ordering is
identical across all three models despite their differing absolute accuracy, indicating
that the difficulty gradient is a property of the benchmark's
constituent sources rather than an artefact of any single model. This
pattern is consistent with per-source vocabulary scale, narrower
predicate and action vocabularies correspond to easier retrieval. Full per-source, per-model details are provided in Sec.~D.4 of the supplementary material.

\subsection{Linear Probe}
 
As an additional benchmark on the SCLARO dataset, we freeze each model
entirely and train task-specific linear heads to predict the action,
object, or relation for a given image, fine-tuned for $6$ epochs.
\cref{tab:lp_ablation} shows that ScenarioCLIP outperforms PyramidCLIP,
with improvements of $3.83\%/4.56\%/3.93\%$ on Top-$1$
\textit{Actions}/\textit{Objects}/\textit{Relations}.
 
\subsection{Predicate \& Scene Graph Classification}
 
To benchmark relation understanding directly, we evaluate predicate
classification (PredCls) and scene graph classification
(SGCls)~\cite{xu2017scenegraph} over the test set of the SCLARO
dataset. In
PredCls, object boxes and categories are given, and each relation is
scored by cosine similarity between the relation-encoder feature (union
of the two object regions) and the text embeddings of all candidate
$(object_1, predicate, object_2)$ triplets. SGCls additionally requires
predicting object categories. We report mean Recall@$K$ (mR@$K$), which
averages Recall@$K$ per predicate (restricted to predicates with sufficient test set coverage, see Sec. D.7 of the supplementary for details) to counter the long-tailed predicate
distribution. As
shown in \cref{tab:pred-sgcls}, ScenarioCLIP improves over PyramidCLIP
by $+1.53/+2.10/+2.27$ on mR@$1/5/10$ (PredCls) and $+1.82/+3.22/+3.16$
(SGCls). 

\begin{table*}[tp]
% \tiny
\centering
\footnotesize
\setlength{\tabcolsep}{3pt}

\resizebox{0.95\textwidth}{!}{%
\begin{tabular}{lccc ccc cccccc}
\toprule
& \multicolumn{3}{c}{\textbf{MS-COCO (Objects)}}
& \multicolumn{3}{c}{\textbf{Visual Genome (Objects)}}
& \multicolumn{6}{c}{\textbf{Visual Genome (Relations)}} \\
\cmidrule(lr){2-4}\cmidrule(lr){5-7}\cmidrule(lr){8-13}
\textbf{Model} & Top-1 & Top-5 & Top-10
               & Top-1 & Top-5 & Top-10
               & Top-1 & Top-5 & Top-10
               & mR@1 & mR@5 & mR@10 \\
\midrule
CLIP~\cite{radford2021learningtransferablevisualmodels}
    & \textbf{31.92} & \textbf{56.96} & \textbf{70.25}
    & 29.43 & \textbf{55.98} & \textbf{66.27}
    & 0.80 & 3.55 & 5.08
    & 1.61 & 4.58 & 8.63 \\
PyramidCLIP~\cite{gao2022pyramidcliphierarchicalfeaturealignment}
    & 23.33 & 45.24 & 56.69
    & 21.94 & 45.52 & 56.17
    & 3.36 & 12.50 & 19.58
    & 4.25 & \textbf{13.04} & 19.07 \\
ScenarioCLIP w/o KD
    & 29.89 & 46.85 & 56.83
    & 29.13 & 49.16 & 56.69
    & 3.40 & 12.22 & 19.52
    & 3.94 & 12.16 & 18.75 \\
ScenarioCLIP
    & 31.31 & 48.74 & 58.74
    & \textbf{29.70} & 50.13 & 57.70
    & \textbf{3.48} & \textbf{13.17} & \textbf{20.18}
    & \textbf{4.30} & 13.00 & \textbf{19.23} \\
\bottomrule
\end{tabular}%
}
\caption{Out-of-domain generalization. Object classification on MS-COCO~\cite{lin2015microsoftcococommonobjects}
 and Visual Genome~\cite{krishna2016visualgenomeconnectinglanguage} , evaluated using plain bbox crops; and relation zero-shot
image$\rightarrow$text retrieval generalization on Visual Genome. 
CLIP~\cite{radford2021learningtransferablevisualmodels} denotes the vanilla CLIP model.%CLIP~\cite{radford2021learningtransferablevisualmodels} denotes the original ViT-B/32
%weights. % without SCLARO Dataset pretraining.
}
\label{tab:external}
\end{table*}

\subsection{Object Detection}
To establish object detection as a benchmark task for the SCLARO 
dataset, we train a Faster R-CNN~\cite{ren2016fasterrcnnrealtimeobject} 
head using our encoders as backbones. For PyramidCLIP, the single visual 
encoder feeds both the RPN and RoI heads. For ScenarioCLIP, the global 
visual encoder provides RPN features and the object encoder supplies 
RoI features. Models are trained for $10$ epochs on the SCLARO dataset. 
The long-tailed nature of the dataset ($4{,}812$ object classes) makes 
this challenging. ScenarioCLIP yields consistent gains over PyramidCLIP: 
$AP^{bb}$ improves from $9.7$ to $9.9$, $AP_{50}^{bb}$ from $18.6$ to 
$18.9$, and $AP_{75}^{bb}$ from $8.9$ to $9.1$ (\cref{tab:det_ap_pct}), 
suggesting disentangled object-level representations are a slightly 
better detection backbone than a single shared encoder.

%To evaluate whether improved relational modelling transfers beyond 
%retrieval, we perform object detection using a Faster 
%R-CNN~\cite{ren2016fasterrcnnrealtimeobject} head with our encoders as 
%backbones. For PyramidCLIP, the single visual encoder feeds both the RPN 
%and RoI heads; for ScenarioCLIP, the global visual encoder provides RPN 
%features and the object visual encoder supplies RoI features. Models are 
%trained for $10$ epochs on the SCLARO Dataset. The long-tailed 
%nature of the dataset ($4{,}812$ object classes) makes this a 
%challenging task. ScenarioCLIP yields consistent gains over 
%PyramidCLIP: $AP^{bb}$ improves from $9.7$ to $9.9$, %$AP_{50}^{bb}$ 
%from $18.6$ to $18.9$, and $AP_{75}^{bb}$ from $8.9$ to $9.1$ 
%(\cref{tab:det_ap_pct}).

\begin{table}[t]
  \centering
  \small
  \setlength{\tabcolsep}{8pt}
  
  \begin{tabular}{lccc}
    \toprule
    \textbf{Model} & \textbf{$AP^{bb}$} & \textbf{$AP_{50}^{bb}$} 
    & \textbf{$AP_{75}^{bb}$} \\
    \midrule
    PyramidCLIP~\cite{gao2022pyramidcliphierarchicalfeaturealignment} 
    & 9.7 & 18.6 & 8.9 \\
    ScenarioCLIP w/o KD & \textbf{9.9} & 18.7 & \textbf{9.1} \\
    ScenarioCLIP        & \textbf{9.9} & \textbf{18.9} & \textbf{9.1} \\
    \bottomrule
  \end{tabular}
  \caption{Object detection results on the SCLARO Dataset.}
  \label{tab:det_ap_pct}
\end{table}

\subsection{Generalisation to External Benchmarks}

To evaluate whether scenario-level pretraining preserves general-purpose 
representations, we evaluate out-of-domain object and relation 
generalisation. For objects, we evaluate classification on MS-COCO 
val2017 ~\cite{lin2015microsoftcococommonobjects} (80 classes) and Visual Genome ~\cite{krishna2016visualgenomeconnectinglanguage} (top-150 vocabulary~\cite{lu2016visualrelationshipdetectionlanguage,zellers2018neuralmotifsscenegraph}) against vanilla CLIP ViT-B/32~\cite{radford2021learningtransferablevisualmodels} (\cref{tab:external}). ScenarioCLIP 
achieves $31.31\%$ Top-1 on COCO and $29.70\%$ on Visual Genome, matching or 
exceeding vanilla CLIP ($31.92\%$ and $29.43\%$), and outperforms PyramidCLIP 
$+7.98\%$ on COCO and $+7.76\%$ on Visual Genome Top-1. This indicates 
SCLARO pretraining does not cause catastrophic forgetting, and that 
ScenarioCLIP's object encoder generalises better than PyramidCLIP's 
shared encoder.

For relations, we evaluate zero-shot image$\rightarrow$text retrieval generalisation 
on Visual Genome~\cite{krishna2016visualgenomeconnectinglanguage} (top-50 predicate vocabulary~\cite{lu2016visualrelationshipdetectionlanguage,zellers2018neuralmotifsscenegraph}), using SAM-derived relation-focused regions 
constructed from VG's object boxes via SCLARO's construction procedure. CLIP \cite{radford2021learningtransferablevisualmodels} is evaluated on the source 
image, lacking a dedicated relation encoder. All three SCLARO-trained models substantially outperform CLIP, with
ScenarioCLIP achieving $3.48\%$ Top-$1$ versus CLIP's $0.80\%$, with similar 
margins at Top-$5/10$ and mR@$K$. This indicates relation-focused-region 
pretraining transfers beyond the dataset's domain, underscoring SCLARO's 
value for learning relational representations.

\subsection{Ablation Study} \label{sec:ablation}
To isolate the contribution of the KD objective as a
regularizer, we compare ScenarioCLIP against a variant trained without KD. \Cref{tab:zs_retrieval_full} shows that KD
provides consistent gains in zero-shot retrieval, particularly for
\textit{Objects} ($+0.60\%$ Top-1) and \textit{Relations} ($+0.11\%$
Top-1). Object detection (\cref{tab:det_ap_pct}) shows a minor
improvement of $0.2$ in $AP^{bb}_{50}$, while linear-probe results
(\cref{tab:lp_ablation}) show marginal differences. On out-of-domain
generalisation (\cref{tab:external}), ScenarioCLIP outperforms
ScenarioCLIP w/o KD consistently, supporting KD's role as a geometric
regulariser that improves transfer to unseen domains over in-domain
classification capacity. We compare this asymmetric intra-modal KD
direction against cross-level and symmetric alternatives in Sec.~D.10 of the
supplementary material.

\section{Conclusion}

% We introduced the SCLARO Dataset, a 615,805-image corpus annotated with
% action captions, object bounding boxes, and relation-focused regions
% across five diverse visual domains, providing a new resource for
% scenario-level VLP and evaluation. To benchmark
% the dataset, we proposed ScenarioCLIP, a hierarchical tri-level
% reference model with disentangled encoders and asymmetric EMA-based
% knowledge distillation. We evaluated PyramidCLIP and ScenarioCLIP across
% a comprehensive suite of tasks, namely zero-shot retrieval, linear
% probe, object detection, predicate classification, scene-graph
% classification, and out-of-domain generalisation. ScenarioCLIP's
% disentangled encoders improve over PyramidCLIP, particularly at the
% object and relation levels and on out-of-domain generalisation, while
% both models retain competitive performance on external benchmarks. We
% hope the SCLARO Dataset serves as a foundation for future research into
% grounded, compositional scene understanding.

We introduced the SCLARO dataset, a 615,805-image corpus annotated with
action captions, object bounding boxes, and relation-focused regions
across five visual domains, as a resource for scenario-level
vision-language pretraining and evaluation. To benchmark it, we
proposed ScenarioCLIP, a tri-level model with disentangled encoders and
asymmetric EMA-based knowledge distillation. Across zero-shot
retrieval, linear probe, object detection, predicate classification,
scene-graph classification, and out-of-domain generalisation,
ScenarioCLIP improves over
PyramidCLIP~\cite{gao2022pyramidcliphierarchicalfeaturealignment}'s
single shared encoder, particularly at the object and relation levels
and on out-of-domain generalisation, while both remain competitive on
external benchmarks. We hope SCLARO serves as a foundation for grounded, compositional scene understanding, including future work on scene graph generation, relational reasoning, and other structured vision-language tasks.
\section*{Limitations}
%The SCLARO Dataset annotations are generated via an automated VLM pipeline. Although manually reviewed annotations identified as incorrect were excluded, some hallucinated objects or imprecise relation triplets may remain undetected at this scale. The relation vocabulary is naturally long-tailed, with many rare predicate expressions appearing in few images. GroundingDINO detection quality bounds the precision of focused regions, and the dataset draws from five specific sources which may introduce domain bias not representative of all real-world scenarios.

% The SCLARO dataset annotations are generated via an automated VLM 
% pipeline. Although incorrect annotations identified during manual 
% review were excluded, some hallucinated objects or imprecise relations 
% may remain undetected at this scale. The relation vocabulary is 
% naturally long-tailed, with many rare predicates appearing in few 
% images. GroundingDINO bounds focused-region precision, and the dataset 
% draws from five sources, which may introduce domain bias 
% unrepresentative of real-world scenarios.
The SCLARO dataset annotations are generated via an automated pipeline. Although flagged incorrect annotations were excluded, some hallucinated objects or imprecise relations may remain undetected at this scale. The relation vocabulary is naturally long-tailed, with many rare predicates appearing in a few images. GroundingDINO bounds focused region precision and the five source datasets may introduce bias unrepresentative of real-world scenarios.

%% file: sec/A_impl_details.tex
% ============================================================
%  Appendix A — Dataset Generation Pipeline
%  Drop this into supplementary.tex as:
%  \input{sec/A_impl_details}
% ============================================================

% \section*{Code and Dataset Availability}

% Code for dataset generation, model training, and evaluation
% experiments described in this paper is provided with this
% supplementary material.

% Annotations for the SCLARO Dataset (action labels, object bounding
% boxes, and relation triplets, across all five constituent sources) are
% provided with this supplementary material as JSON metadata files. The
% complete dataset, including relation-focused region images, will be
% made publicly available by the camera-ready deadline.
\section{Dataset Generation Pipeline}
\label{sec:pipeline_details}

\subsection{Stage-1: Structured Scene Annotation}
\label{sec:prompt_design}

Stage-1 of the annotation pipeline uses Ovis-Gemma 9B~\cite{gemmateam2024gemma2improvingopen,lu2024ovis}
to generate a global action caption, a list of unique objects, and a
set of relation triplets for each image. The model is queried with the
following prompt:
 
\begin{quote}
\ttfamily
Given the image, return only the following information:\\
1. A list of major unique objects in the image (limit to 10 objects).\\
2. The relations between those objects in the form of relation
triplets: $\langle$object1$\rangle$ $\langle$relation$\rangle$
$\langle$object2$\rangle$.\\
3. Action should be a single word of a broad class, like "playing",
"eating", "running", "talking", "driving" etc. Avoid specific actions
like "serving", "gulping", "steering" etc.\\
4. The description tag should be one or two sentences long, describing
the image in general terms.\\
Do not provide any extra information or descriptions.\\
Give the output in a json format, example:\\
\{\\
\ \ "action": "<some action>"\\
\ \ "objects": ["object1", "object2", ...],\\
\ \ "relations": [\\
\ \ \ \ ["object1", "relation", "object2"],\\
\ \ \ \ ...\\
\ \ ]\\
\ \ "dense caption": "The image shows a... <description>"\\
\}
\end{quote}
 
We experimented with several prompt formulations during development,
adjusting syntax, output structure, and the level of guidance provided
for relation detection. The prompt above was selected based on output
consistency, specifically its tendency to produce well-formed JSON with
correctly structured relation triplets and minimal hallucinated
objects. Ovis-Gemma 9B \cite{gemmateam2024gemma2improvingopen,lu2024ovis} was chosen as the Stage-1 VLM for its strong
instruction-following capability on structured generation tasks with
visual inputs.

\subsection{Stage-2: Object Localisation}
\label{sec:grounding_dino_details}

Stage-2 uses GroundingDINO~\cite{liu2023grounding} (\texttt{grounding-dino-base}) to predict bounding boxes for the objects identified in Stage-1. The object list from Stage-1 is concatenated into a single period-separated, lowercased string and passed as the text prompt to GroundingDINO. Detection uses a box confidence threshold of 0.25 and a text-matching threshold of 0.25. Non-maximum suppression (NMS) is applied post-hoc with an IoU threshold of 0.4 to remove duplicate detections. Images for which the VLM produced empty object lists or empty relation sets are excluded from the dataset.

\subsection{Stage-3: Focused Region Construction}
\label{sec:focused_region_details}

Stage-3 constructs a spatially grounded focused region for each relation triplet $(object_1, relation, object_2)$. Given the bounding boxes from Stage-2, SAM~\cite{kirillov2023segment} (\texttt{sam-vit-huge}) is used to generate per-object segmentation masks. For each relation triplet, we identify the closest pair of $object_1$ and $object_2$ instances by Euclidean distance between bounding-box centres, and retrieve their corresponding masks.

Let $\mathbf{c}_1$ and $\mathbf{c}_2$ denote the centres of mass of the two object masks. We construct an RBF-weighted mask for each object:
\begin{equation}
    \text{RBF}_k(\mathbf{x}) = e^{-\frac{\|\mathbf{x} - \mathbf{c}_k\|^2}{2\sigma^2}}, \quad k \in \{1, 2\},
\end{equation}
with $\sigma = 100$ pixels. These are combined using distance-based weights that interpolate smoothly between the two objects:
\begin{equation}
    w_1(\mathbf{x}) = \frac{d_2(\mathbf{x})}{d_1(\mathbf{x}) + d_2(\mathbf{x})}, \quad
    w_2(\mathbf{x}) = \frac{d_1(\mathbf{x})}{d_1(\mathbf{x}) + d_2(\mathbf{x})},
\end{equation}
where $d_k(\mathbf{x}) = \|\mathbf{x} - \mathbf{c}_k\|$ is the Euclidean distance from pixel $\mathbf{x}$ to $\mathbf{c}_k$. The blended focused region mask is then:
\begin{equation}
    M(\mathbf{x}) = \text{RBF}_1(\mathbf{x}) \cdot w_1(\mathbf{x}) + \text{RBF}_2(\mathbf{x}) \cdot w_2(\mathbf{x}).
\end{equation}
Pixels within the union of the two object masks are multiplied by $M(\mathbf{x})$ to produce a relation-highlighted foreground. Pixels outside the union are replaced by a Gaussian-blurred version of the original image (kernel $\sigma = 2$):
\begin{equation}
    G(x, y) = \frac{1}{2\pi\sigma^2} e^{-\frac{x^2 + y^2}{2\sigma^2}},
\end{equation}
yielding a focused region image that emphasises the spatial interaction between $object_1$ and $object_2$ while preserving ambient scene context.

%% file: sec/B_data_curation.tex
% ============================================================
%  Appendix B — Data Curation
%  Drop into supplementary.tex as:
%  \input{sec/B_curation}
% ============================================================

\section{Data Curation}
\label{sec:curation}

\subsection{Action Classes}

The five constituent sources differ in how action labels are obtained.
For Multi-Moments in Time~\cite{monfort2021multimomentstimelearninginterpreting}, ground-truth action class labels are provided directly and reused without modification.
For CC3M~\cite{sharma-etal-2018-conceptual}, OpenPSG~\cite{zhou2024openpsgopensetpanopticscene}, Panoptic Video Scene Graph (PVSG)~\cite{yang2023panopticvideoscenegraph}, and KITTI2015~\cite{geiger2012cvpr}, action labels are generated by the Stage-1 VLM pipeline described in \cref{sec:pipeline_details}.

\begin{figure}[h]
  \centering
  \includegraphics[width=0.4\textwidth]{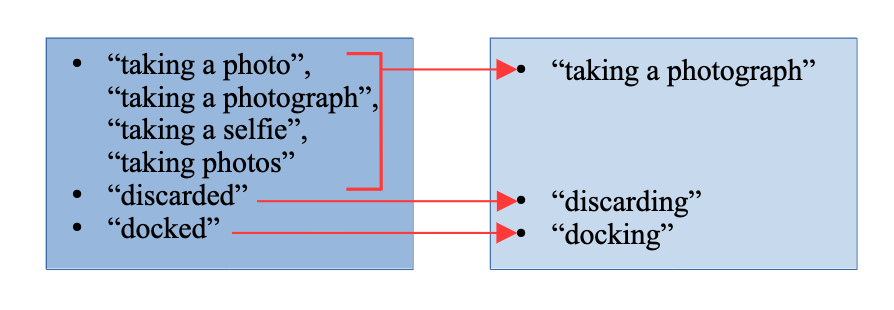}
  \caption{\textbf{Action processing and consolidation.} Diverse raw action phrases are normalised and merged into a smaller set of canonical action labels.}
  \label{fig:actcure}
\end{figure}

VLM-generated action phrases are normalised and consolidated into a reduced set of canonical action classes.
Diverse surface forms referring to the same activity are merged into a single label, for instance, \textit{taking a photo}, \textit{taking a photograph}, \textit{taking a selfie}, and \textit{taking photos} are all consolidated into \textit{taking a photograph} (\cref{fig:actcure}).
Images for which the VLM assigned no action label (0.1\% of the dataset, $n = 568$) are retained since their object and relation annotations remain valid.

\subsection{Object Classes}

The raw object lists produced by the VLM pipeline contain several types of noise: semantically related objects listed as distinct entries, multiple unrelated objects grouped into a single bounding box label, adjectives and colour descriptors prepended to object names, and both singular and plural forms of the same object appearing separately.

\begin{figure*}[t]
  \centering
  \includegraphics[width=\textwidth]{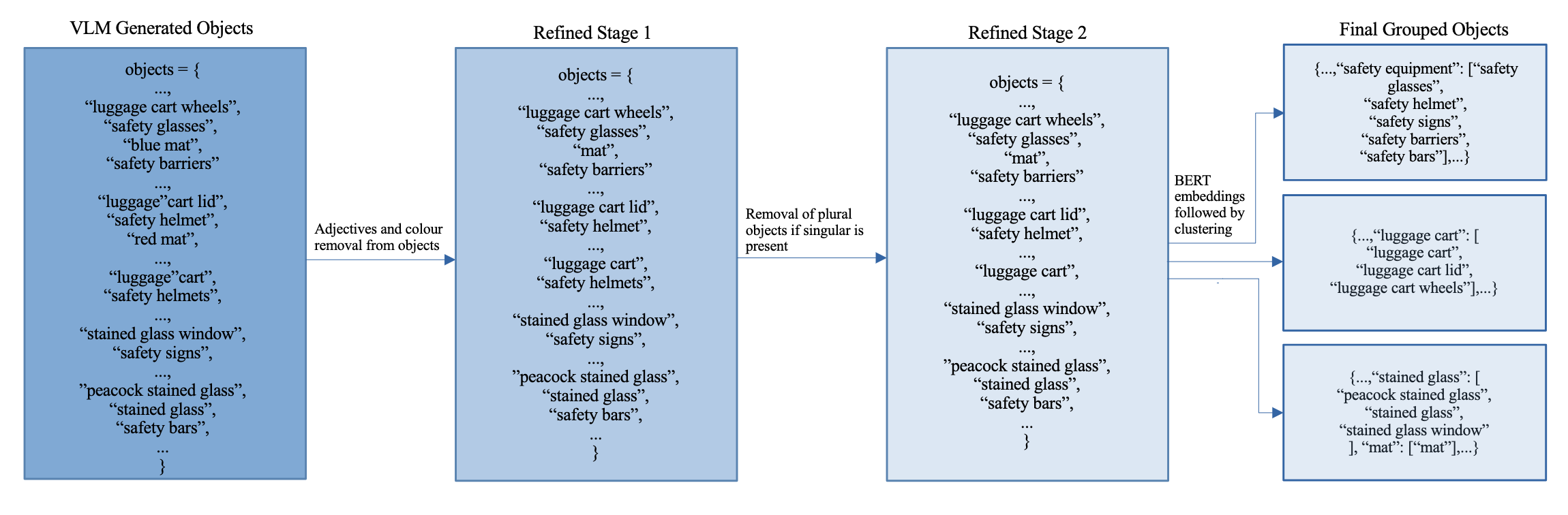}
  \caption{\textbf{Object processing pipeline.} Raw VLM-generated object strings are refined into final grouped object classes by removing adjectives and colour descriptors, eliminating redundant plural forms, and clustering semantically similar objects using BERT embeddings and HDBSCAN.}
  \label{fig:objcure}
\end{figure*}

We address these issues through a three-step filtering process.
First, we reduce the raw object list by taking the intersection of objects identified across all VLM stages for each sample, and the union of these intersections across all samples, discarding labels that could not be consistently identified.
Second, we standardise object names by removing colour descriptors (using a predefined colour term list) and adjectives (using the NLTK~\cite{bird2009natural} POS tagger), and by removing plural forms when both singular and plural variants are present.
Third, to group semantically related objects, we embed all object names using BERT~\cite{devlin2019bertpretrainingdeepbidirectional} and cluster the embeddings using HDBSCAN~\cite{10.1007/978-3-642-37456-2_14}, with cosine distance and mutual reachability:
\[
d_{\text{mreach-}k}(a, b) = \max \{ \text{core}_k(a),\, \text{core}_k(b),\, d(a, b) \}.
\]
Each resulting cluster is manually assigned a semantically meaningful label, for example, the cluster containing \textit{garbage truck}, \textit{trash can}, \textit{waste bin}, \textit{trash bag}, and \textit{garbage bags} is labelled \textit{waste management}.
The full object processing pipeline is illustrated in \cref{fig:objcure}.

\begin{figure*}[t]
  \centering
  \includegraphics[width=0.7\textwidth]{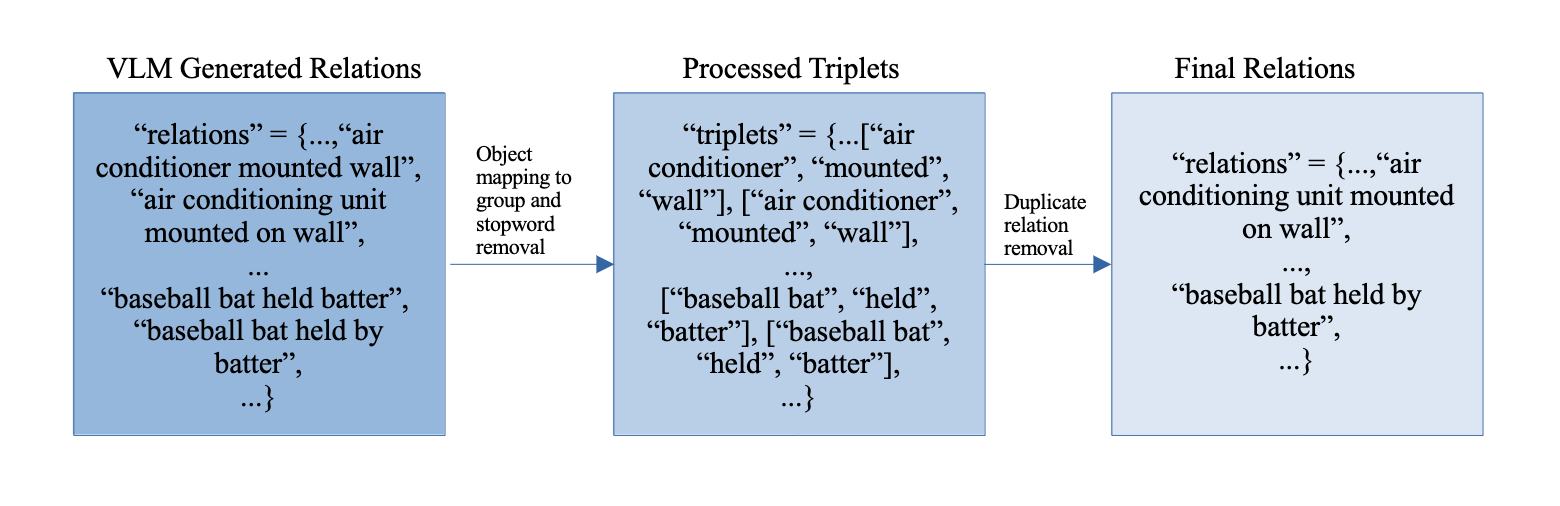}
  \caption{\textbf{Relation processing pipeline.} Raw VLM-generated relation strings are converted into relation triplets by mapping objects to their canonical group labels, removing stopwords, and collapsing duplicate triplets into a final relation set.}
  \label{fig:relcure}
\end{figure*}

\subsection{Relation Classes}

The VLM pipeline frequently generates semantically equivalent relation strings that differ only in stopwords, for example, \textit{air conditioner mounted wall} and \textit{air conditioner mounted on wall} carry identical meaning.
To consolidate the relation vocabulary, we first extract the structured triplet $(object_1, relation, object_2)$ from each generated relation string.
We then replace $object_1$ and $object_2$ with their canonical cluster labels from the object curation step above, and remove stopwords from the relation phrase.
If the resulting normalised triplet duplicates an existing entry, one instance is discarded (\cref{fig:relcure}).
The relation vocabulary is further refined by a final pass of manual curation. Within a single image, the same canonical object label may appear multiple times as distinct instances, each with its own bounding box (e.g. two separate \emph{person} detections at different image locations), and the same canonical relation triplet text may correspond to multiple distinct instances when different object pairs share the same predicate (e.g. two separate \emph{person holding cup} interactions involving different people and cups). In both cases, each instance is retained as a separate entry rather than merged. The final action, object, and relation label spaces are deduplicated across all five constituent sources, so that the same canonical label or relation triplet refers to a single shared class regardless of which source dataset it originated from.

\section{Qualitative Annotation Examples}
\label{sec:qualitative_examples}
\begin{figure*}[t]
    \centering
    \includegraphics[width=\textwidth]{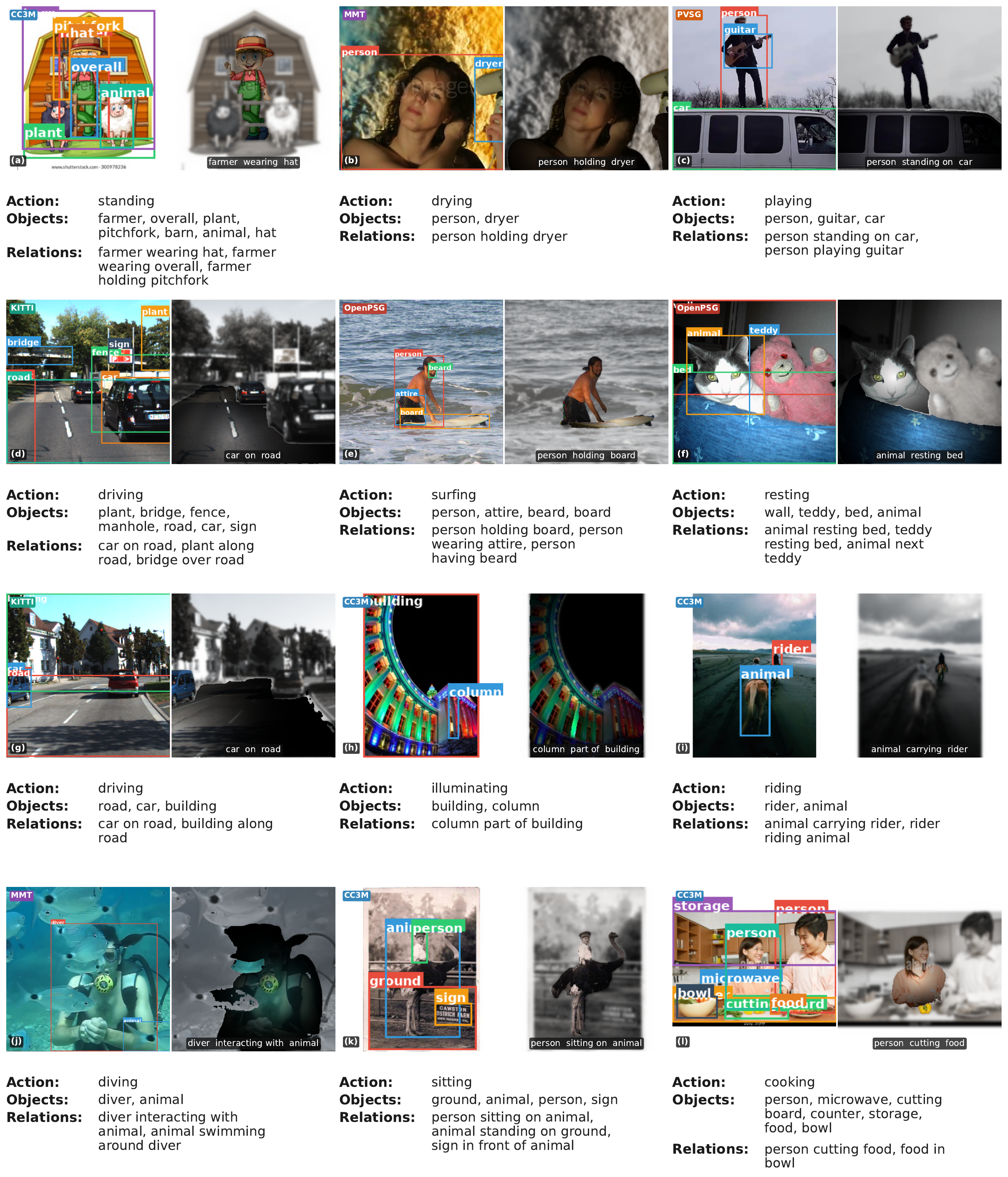}
    \caption{\textbf{Qualitative annotation examples from the SCLARO dataset.} Each panel shows the annotated image with object bounding boxes (left) and a relation-focused region (right).}
    \label{fig:annotation_quality}
\end{figure*}

\Cref{fig:annotation_quality} shows twelve representative
examples of correctly annotated images, spanning all five constituent
source datasets (CC3M~\cite{sharma-etal-2018-conceptual}, MMT~\cite{monfort2021multimomentstimelearninginterpreting}, PVSG~\cite{yang2023panopticvideoscenegraph}, OpenPSG~\cite{zhou2024openpsgopensetpanopticscene}, and KITTI2015~\cite{geiger2012cvpr}). Each
panel displays the source image with predicted object bounding boxes
(left) and one relation-focused region with its corresponding triplet
overlaid (right), alongside the full action caption, object list, and
relation triplets. These examples illustrate the range of scenes the
pipeline handles, from posed studio photographs and driving scenes to
informal video frames, and the diversity of actions, objects, and
relations captured across the dataset's five constituent sources.

%% file: sec/C_experiment_details.tex
% ============================================================
%  Appendix C — Experiment Details
%  Drop this into supplementary.tex as:
%  \input{sec/C_experiment_details}
% ============================================================
%\begin{table*}[h]
%\centering
%\small
%\caption{Predicate classification (PredCls) and scene graph classification (SGCls) results, reporting mean Recall@$K$ (mR@$K$) over the 1,320-predicate evaluation vocabulary ($\geq$5 test instances, 98.9\% coverage).}
%\begin{tabular}{l ccc ccc}
%\toprule
%& \multicolumn{3}{c}{PredCls} & \multicolumn{3}{c}{SGCls} \\
%\cmidrule(lr){2-4}\cmidrule(lr){5-7}
%\textbf{Model} & mR@1 & mR@5 & mR@10 & mR@1 & mR@5 & mR@10 \\
%\midrule
%PyramidCLIP~\cite{gao2022pyramidcliphierarchicalfeaturealignment} & 7.52 & 20.57 & 28.33 & 6.56 & 18.01 & 25.57 \\
%ScenarioCLIP w/o KD & 9.01 & \textbf{22.81} & \textbf{31.16} & 8.18 & 20.92 & 28.65 \\
%ScenarioCLIP & \textbf{9.05} & 22.67 & 30.60 & \textbf{8.38} & \textbf{21.23} & \textbf{28.73} \\
%\bottomrule
%\end{tabular}
%\label{tab:pred-sgcls}
%\end{table*}

% ============================================================
%  Appendix C — Experiment Details
%  Drop this into supplementary.tex as:
%  \input{sec/C_experiment_details}
% ============================================================

\section{Experiment Details}
\label{experiments_appendix}

\subsection{Pre-training}
\label{sec:pretraining_details}

All ScenarioCLIP variants are pretrained on the SCLARO dataset for 12 epochs using the AdamW optimiser~\cite{loshchilov2019decoupledweightdecayregularization} with a base learning rate of $2\times 10^{-5}$, $\beta = (0.9, 0.999)$, $\varepsilon = 10^{-8}$, and weight decay $0.2$. We employ a linear warm-up over the first 10\% of optimisation steps, where the learning rate increases from $10^{-3}$ of the base value to the full learning rate, followed by cosine annealing to zero.

For ScenarioCLIP, the learnable distillation temperature is optimised as a separate parameter group at $0.1\times$ the base learning rate with no weight decay, which we found to stabilise training. The EMA teacher is maintained with decay $0.9995$ and a 2000-step warm-up before EMA updates begin.

\subsection{Model Capacity and Compute}
\label{sec:capacity}
\Cref{tab:capacity} reports parameter counts and FLOPs for
PyramidCLIP \cite{gao2022pyramidcliphierarchicalfeaturealignment} and ScenarioCLIP. All encoders use the CLIP ViT-B/32 \cite{radford2021learningtransferablevisualmodels}
backbone, so a single image-text encoder pair has identical cost across
models: 151.28M parameters and 14.55 GFLOPs per (image, caption) forward
pass (batch size $1$). PyramidCLIP \cite{gao2022pyramidcliphierarchicalfeaturealignment} shares one encoder
pair across its hierarchy, whereas ScenarioCLIP uses three disentangled
pairs (global, object, relation), giving $3\times$ the parameters
($453.83$M vs.\ $151.28$M).
Crucially, the per-pair forward cost is unchanged: each level is a
standard CLIP pair. At inference, a downstream task invokes only the
relevant pair (e.g.\ object retrieval uses the object encoders alone), so
per-task inference cost matches a single CLIP pair. The additional
parameters reflect \emph{level specialisation} rather than wider or
deeper encoders. During training, a full multi-level forward processes
the global image together with all object crops and relation regions,
costing approximately $(1 + n_O + n_R)\times 14.55$ GFLOPs for an image
with $n_O$ objects and $n_R$ relations.

ScenarioCLIP is pretrained on a single NVIDIA H100 GPU with a batch size
of $128$, for $12$ epochs over the SCLARO dataset, taking approximately
$4$ days of wall-clock time.

\begin{table}[t]
\centering
\small
\begin{tabular}{lccc}
\toprule
\textbf{Model} & \textbf{Enc.\ pairs} & \textbf{Params} & \textbf{GFLOPs/pair} \\
\midrule
CLIP ViT-B/32~\cite{radford2021learningtransferablevisualmodels} & 1 & 151.28M & 14.55 \\
PyramidCLIP~\cite{gao2022pyramidcliphierarchicalfeaturealignment}   & 1 (shared) & 151.28M & 14.55 \\
ScenarioCLIP (ours) & 3 & 453.83M & 14.55 \\
\bottomrule
\end{tabular}
\caption{Parameter and FLOP comparison. A single CLIP ViT-B/32 \cite{radford2021learningtransferablevisualmodels}
image-text encoder pair costs 151.28M parameters and 14.55 GFLOPs per
forward (batch size $=1$). ScenarioCLIP uses three such pairs;
per-pair cost is identical to PyramidCLIP \cite{gao2022pyramidcliphierarchicalfeaturealignment}.}
\label{tab:capacity}
\end{table}

\subsection{Knowledge Distillation Mechanics}
\label{sec:kd_mechanics}
Each per-level distillation term converts embeddings into probability
distributions via a softmax over the embedding's feature dimension,
following standard knowledge distillation
practice~\cite{hinton2015distillingknowledgeneuralnetwork}. Given a
fixed teacher embedding $z_t \in \mathbb{R}^d$ and a set of $N$ student
embeddings $\{z_s^{(i)}\}_{i=1}^N \in \mathbb{R}^d$ (e.g.\ the EMA
global visual embedding teaching the $N_O$ object and $N_R$ relation
visual embeddings of one image), the distillation loss is:
\begin{equation}
\mathcal{L}_{\text{KD}}(z_t, \{z_s^{(i)}\}) =
\frac{T_s^2}{N} \sum_{i=1}^{N}
D_{KL}\Big(
\sigma\big(z_t / T_t\big)
\,\big\|\,
\sigma\big(z_s^{(i)} / T_s\big)
\Big),
\end{equation}
where $\sigma(\cdot)$ denotes softmax over the embedding's feature
dimension, $T_t = 0.07$ is a fixed teacher temperature, and $T_s$ is a
learnable student temperature, parameterised as $T_s = 1 / \exp(\ell)$
with $\exp(\ell)$ clamped to $[1, 100]$ (i.e.\ $T_s \in [0.01, 1.0]$),
mirroring CLIP's learnable logit-scale
parameterisation~\cite{radford2021learningtransferablevisualmodels}.
The $T_s^2$ factor rescales gradients to compensate for their
attenuation at higher temperatures,
following~\cite{hinton2015distillingknowledgeneuralnetwork}. For the
reverse direction, a single student embedding $z_s$ distilled from a
group of $N$ teacher embeddings $\{z_t^{(i)}\}$ (e.g.\ object/relation
text embeddings teaching the global text embedding), the loss is
symmetric in its construction, with the roles of $\sigma(z_s/T_s)$ and
$\sigma(z_t^{(i)}/T_t)$ as the KL divergence's two arguments swapped
accordingly, and the same $T_s, T_t$ used throughout. Teacher embeddings
are always detached from the computational graph (no gradient flows
through $z_t$). In Eq.~2 of the main paper, $D_{KL}(V_G^{(t)} \|
V_{O,i})$ and $D_{KL}(V_G^{(t)} \| V_{R,j})$ instantiate the first
(one-teacher, many-students) form per image, while $D_{KL}(T_{O,i}^{(t)}
\| T_G)$ and $D_{KL}(T_{R,j}^{(t)} \| T_G)$ instantiate the second
(many-teachers, one-student) form.

\begin{table*}[h]
\centering
\small
\setlength{\tabcolsep}{4pt}
\begin{tabular}{llccc|ccc|ccc}
\toprule
& & \multicolumn{3}{c}{PyramidCLIP} & \multicolumn{3}{c}{ScenarioCLIP w/o KD} & \multicolumn{3}{c}{ScenarioCLIP} \\
\cmidrule(lr){3-5}\cmidrule(lr){6-8}\cmidrule(lr){9-11}
& \textbf{Source} & Top-1 & Top-5 & Top-10 & Top-1 & Top-5 & Top-10 & Top-1 & Top-5 & Top-10 \\
\midrule
\multirow{5}{*}{\rotatebox[origin=c]{90}{Action}}
& CC3M \cite{sharma-etal-2018-conceptual}          & 56.37 & 73.36 & 78.84 & 56.38 & 74.35 & 79.68 & 56.66 & 74.29 & 79.31 \\
& MMT \cite{monfort2021multimomentstimelearninginterpreting}            & 42.10 & 62.22 & 69.12 & 48.67 & 69.24 & 75.43 & 48.55 & 68.93 & 75.12 \\
& OpenPSG \cite{zhou2024openpsgopensetpanopticscene}      & 65.75 & 81.79 & 86.05 & 66.37 & 81.21 & 85.23 & 66.88 & 81.22 & 85.01 \\
& PVSG \cite{yang2023panopticvideoscenegraph}          & 82.45 & 93.65 & 95.36 & 83.60 & 94.09 & 95.52 & 83.33 & 93.15 & 94.94 \\
& KITTI2015 \cite{geiger2012cvpr} & 98.90 & 99.51 & 99.69 & 98.71 & 99.26 & 99.33 & 98.65 & 99.20 & 99.33 \\
\midrule
\multirow{5}{*}{\rotatebox[origin=c]{90}{Object}}
& CC3M \cite{sharma-etal-2018-conceptual}          & 34.88 & 59.20 & 66.73 & 48.72 & 64.54 & 69.24 & 49.59 & 64.60 & 69.27 \\
& MMT \cite{monfort2021multimomentstimelearninginterpreting}           & 28.90 & 51.19 & 58.14 & 43.84 & 57.00 & 61.21 & 44.31 & 57.13 & 61.23 \\
& OpenPSG \cite{zhou2024openpsgopensetpanopticscene}       & 37.65 & 59.49 & 65.47 & 48.87 & 62.52 & 66.63 & 49.27 & 62.47 & 66.49 \\
& PVSG \cite{yang2023panopticvideoscenegraph}          & 69.90 & 83.32 & 84.90 & 82.01 & 85.95 & 86.79 & 82.37 & 85.95 & 86.67 \\
& KITTI2015 \cite{geiger2012cvpr} & 49.91 & 65.55 & 66.92 & 60.21 & 65.89 & 66.87 & 60.02 & 65.89 & 66.80 \\
\midrule
\multirow{5}{*}{\rotatebox[origin=c]{90}{Relation}}
& CC3M \cite{sharma-etal-2018-conceptual}          & 9.20  & 27.75 & 38.64 & 11.81 & 30.33 & 40.14 & 11.84 & 30.39 & 40.20 \\
& MMT \cite{monfort2021multimomentstimelearninginterpreting}           & 8.25  & 24.00 & 33.35 & 10.73 & 26.71 & 35.14 & 10.66 & 26.71 & 35.12 \\
& OpenPSG \cite{zhou2024openpsgopensetpanopticscene}       & 11.06 & 31.89 & 44.27 & 14.19 & 34.45 & 45.30 & 14.08 & 34.71 & 45.70 \\
& PVSG \cite{yang2023panopticvideoscenegraph}          & 83.15 & 94.45 & 96.01 & 88.35 & 96.69 & 97.44 & 89.62 & 96.76 & 97.53 \\
& KITTI2015 \cite{geiger2012cvpr} & 25.26 & 59.69 & 72.31 & 33.44 & 67.32 & 77.91 & 34.72 & 67.62 & 78.00 \\
\bottomrule
\end{tabular}
\caption{Per-source zero-shot retrieval accuracy across Action, Object, and Relation tasks, for all three evaluated models.}
\label{tab:per_source_breakdown}
\end{table*}

\subsection{Per-Source Difficulty Breakdown}
\label{sec:per_source_breakdown}
 
% \Cref{tab:per_source_breakdown} reports zero-shot retrieval accuracy broken down by constituent source dataset, for all three evaluated models. Across all three tasks, the relative difficulty ordering of the five sources is identical for PyramidCLIP \cite{gao2022pyramidcliphierarchicalfeaturealignment}, ScenarioCLIP w/o KD, and ScenarioCLIP, despite substantial differences in absolute accuracy between models. Multi-Moments in Time \cite{monfort2021multimomentstimelearninginterpreting} and CC3M \cite{sharma-etal-2018-conceptual} are consistently the most challenging sources, while PVSG \cite{yang2023panopticvideoscenegraph} and KITTI2015 \cite{geiger2012cvpr} are consistently the easiest. This consistency across architecturally distinct models indicates that the observed difficulty gradient is a property of the dataset's constituent sources rather than an artifact of any single model. 
% \Cref{tab:per_source_vocab} reports the underlying vocabulary scale per source, computed over the full dataset: PVSG \cite{yang2023panopticvideoscenegraph} contains only 714 unique relation triplets (versus 129,636 for CC3M \cite{sharma-etal-2018-conceptual} and 105,520 for Multi-Moments in Time \cite{monfort2021multimomentstimelearninginterpreting}), and KITTI2015~\cite{geiger2012cvpr} contains only 5 unique action labels (versus 712 for CC3M \cite{sharma-etal-2018-conceptual}), reflecting the narrower range of actions inherent to outdoor driving footage, directly accounting for why these two sources are consistently the easiest for relation and action retrieval respectively.

\Cref{tab:per_source_breakdown} reports zero-shot retrieval accuracy broken down by constituent source dataset, for all three evaluated models. Across all three tasks, the relative difficulty ordering of the five sources is identical for PyramidCLIP \cite{gao2022pyramidcliphierarchicalfeaturealignment}, ScenarioCLIP w/o KD, and ScenarioCLIP. Multi-Moments in Time \cite{monfort2021multimomentstimelearninginterpreting} and CC3M \cite{sharma-etal-2018-conceptual} are consistently the most challenging sources, while PVSG \cite{yang2023panopticvideoscenegraph} and KITTI2015 \cite{geiger2012cvpr} are consistently the easiest. This consistency across architecturally distinct models indicates that the observed difficulty gradient is a property of the dataset's constituent sources rather than an artifact of any single model.
\Cref{tab:per_source_vocab} reports the underlying vocabulary scale per source, computed over the full dataset: PVSG \cite{yang2023panopticvideoscenegraph} contains only 714 unique relation triplets (versus 129,636 for CC3M \cite{sharma-etal-2018-conceptual} and 105,520 for Multi-Moments in Time \cite{monfort2021multimomentstimelearninginterpreting}), and KITTI2015~\cite{geiger2012cvpr} contains only 5 unique action labels (versus 712 for CC3M \cite{sharma-etal-2018-conceptual}), reflecting the narrower range of actions inherent to outdoor driving footage. This difficulty gradient reflects training-time exposure density rather than vocabulary size alone: a source with a small, concentrated vocabulary repeats the same labels far more often during pretraining (e.g. PVSG's 54,223 images cover only 714 triplets, an average of $\sim$76 occurrences per triplet, versus CC3M's $\sim$2 occurrences per triplet across 264,838 images and 129,636 triplets), giving the model substantially more supervision per label and a correspondingly easier retrieval target at test time.

% \Cref{tab:per_source_vocab} reports the underlying vocabulary scale per source, computed over the full dataset: PVSG \cite{yang2023panopticvideoscenegraph} contains only 57 unique predicates (versus 5,814 for CC3M \cite{sharma-etal-2018-conceptual} and 5,198 for Multi-Moments in Time \cite{monfort2021multimomentstimelearninginterpreting}), and KITTI2015 \cite{geiger2012cvpr} contains only 5 unique action labels (versus 712 for CC3M \cite{sharma-etal-2018-conceptual}) reflecting the narrower range of actions inherent to outdoor driving footage, directly accounting for why these two sources are consistently the easiest for relation and action retrieval respectively.

\begin{table}[h]
\centering
\small
\begin{tabular}{lccccc}
\toprule
\textbf{Source} & Images & \#Actions & \#Objects & \#Relations \\
\midrule
CC3M           & 264{,}838 & 712 & 3{,}478 & 129{,}636 \\
MMT            & 226{,}066 & 109 & 3{,}453 & 105{,}520 \\
OpenPSG        & 62{,}685  & 310 & 1{,}781 & 45{,}874 \\
PVSG           & 54{,}223  & 113 & 96      & 714 \\
KITTI2015      & 7{,}993   & 5   & 121     & 1{,}431 \\
\bottomrule
\end{tabular}
\caption{Vocabulary scale per source, computed over the full dataset. \#Actions, \#Objects, and \#Triplets count the number of distinct (deduplicated) labels or triplets appearing within each source. The dataset-level totals of 740 action classes, 4{,}812 object classes, and 225{,}609 relation triplets are obtained by deduplicating across all five sources.}
\label{tab:per_source_vocab}
\end{table}

\subsection{Linear Probe}
\label{sec:lp_details}

For linear probing, all backbone parameters are frozen and a single linear layer of size $512 \rightarrow C$ (where $C$ is the number of classes for the task) is trained for 6 epochs using AdamW~\cite{loshchilov2019decoupledweightdecayregularization} with learning rate $2\times 10^{-5}$ and cross-entropy loss over the SCLARO labels.

\subsection{Object Detection}
\label{sec:det_details}

Object detection uses Faster R-CNN~\cite{ren2016fasterrcnnrealtimeobject} with our pretrained encoders as backbones. PyramidCLIP~\cite{gao2022pyramidcliphierarchicalfeaturealignment} uses a single unified visual encoder for both the RPN and RoI heads. ScenarioCLIP employs separate encoders: the global visual encoder provides RPN features for proposal generation, and the object visual encoder supplies RoI features for classification and regression. In all cases, backbone features are converted into multi-scale inputs via a lightweight convolutional pyramid followed by an FPN.

The detector is trained for 10 epochs using AdamW~\cite{loshchilov2019decoupledweightdecayregularization} with a backbone learning rate of $1\times 10^{-5}$ and larger rates for the FPN and detection heads, with 5\% linear warm-up followed by cosine decay. To handle the pronounced long-tailed distribution (4,812 object classes), we adopt an Instance-Aware Sampler~\cite{yaman2023instanceawarerepeatfactorsampling} that assigns sampling weights inversely to per-class frequency, and apply class-targeted Copy-Paste augmentation for categories with fewer than 50 training instances. At inference, we set box score threshold $10^{-4}$, NMS at 0.5, and allow up to 300 detections per image.

\begin{table*}[h]
\centering
\small
\begin{tabular}{l ccc ccc}
\toprule
& \multicolumn{3}{c}{Frozen Encoder} & \multicolumn{3}{c}{Trainable Encoder} \\
\cmidrule(lr){2-4}\cmidrule(lr){5-7}
\textbf{Model} & Dice ($\uparrow$) & IoU ($\uparrow$) & MAE ($\downarrow$) & Dice ($\uparrow$) & IoU ($\uparrow$) & MAE ($\downarrow$) \\
\midrule
PyramidCLIP~\cite{gao2022pyramidcliphierarchicalfeaturealignment} & \textbf{0.6378} & \textbf{0.5269} & 0.1535 & 0.5971 & 0.4858 & 0.1625 \\
ScenarioCLIP w/o KD & 0.6340 & 0.5242 & 0.1512 & \textbf{0.6085} & \textbf{0.4962} & 0.1643 \\
ScenarioCLIP & 0.6230 & 0.5145 & \textbf{0.1509} & 0.6010 & 0.4892 & \textbf{0.1636} \\
\bottomrule
\end{tabular}
\caption{Relation localisation performance for frozen vs.\ trainable visual encoders.}
\label{tab:rel_localization}
\end{table*}

\subsection{Predicate Classification \& Scene Graph Classification}
\label{sec:predcls_details}

This section provides full protocol detail for the predicate
classification (PredCls) and scene graph classification (SGCls)~\cite{xu2017scenegraph} results
reported in the main paper.
 
For predicate classification (PredCls), we use ground-truth bounding
boxes and categories of the two objects involved in each annotated
relation, denoted $object_1$ and $object_2$. We build a predicate
vocabulary $P$ of $1{,}320$ predicates, restricting to predicates
appearing in at least $5$ test instances. We exclude rarer predicates
because per-predicate recall computed over fewer than $5$ instances is
dominated by single-instance variance rather than informative model
behavior, and would inject noise rather than signal into mR@$K$. This
threshold retains $99\%$ of test relation instances. At test time,
for each relation we extract a visual feature from the union of the
$object_1$ and $object_2$ regions using the relation visual encoder, and
construct text features for all candidate triplets $(object_1,
predicate, object_2)$ with $predicate \in P$ using the relation text
encoder. Predicate scores are cosine similarities between the visual and
text features. We report mean Recall@$K$ (mR@$K$, averaged per-predicate
Recall@$K$) for $K \in \{1, 5, 10\}$, a standard metric for evaluating
long-tailed predicate distributions in scene graph
generation~\cite{chen2019kern,tang2019vctree}.
 
For scene graph classification (SGCls), object categories are also
predicted. We build an object vocabulary $C$ from the training split and
classify each object crop by comparing its visual embedding against
text embeddings for all $c \in C$. These predicted labels replace
ground-truth categories when forming relation prompts $(object_1,
predicate, object_2)$. Predicates are then scored and ranked as above.
 
Unlike the in-domain zero-shot retrieval and detection results reported
in the main paper, the contribution of knowledge distillation is mixed
at this finer-grained evaluation: in PredCls, ScenarioCLIP without KD
attains marginally higher mR@5 and mR@10 ($22.81/31.16$ vs.\
$22.67/30.60$), while ScenarioCLIP with KD remains ahead on mR@1
($9.05$ vs.\ $9.01$). In SGCls, ScenarioCLIP with KD is consistently
ahead across all three cutoffs. We attribute this to the long-tailed
predicate vocabulary used for this evaluation ($1{,}320$ predicates, $\geq 5$ test instances): KD's benefit on global representation
structure does not uniformly transfer to fine-grained predicate
discrimination under PredCls, whereas the joint demand of correct
object classification and relation scoring in SGCls benefits more
consistently from the stabilised embedding space that KD provides.

\begin{figure*}[t]
    \centering
    \includegraphics[width=\textwidth]{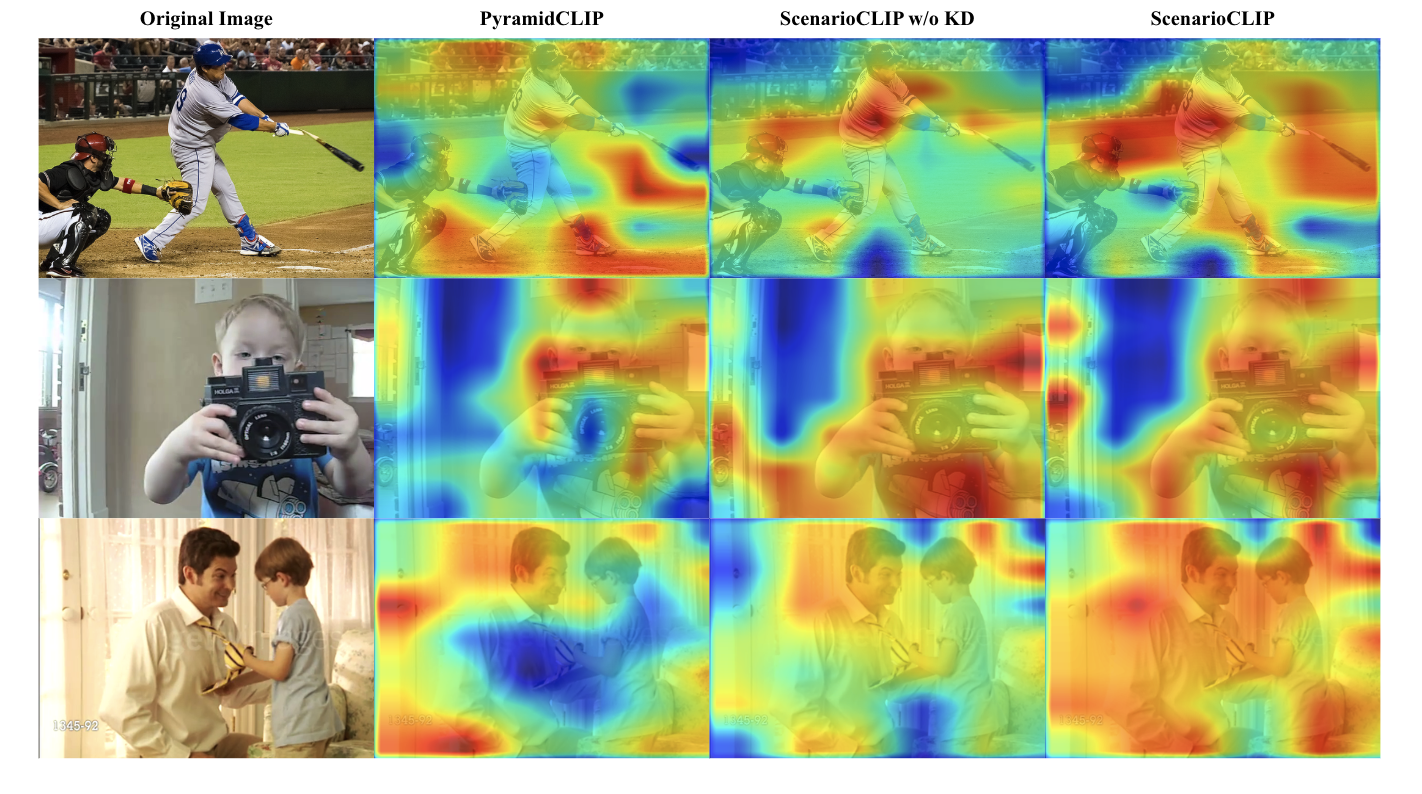}
    \caption{\textbf{Grad-CAM~\cite{Selvaraju_2019} visualisations} of the object encoder for PyramidCLIP~\cite{gao2022pyramidcliphierarchicalfeaturealignment}, ScenarioCLIP w/o KD, and ScenarioCLIP. ScenarioCLIP produces more focused and semantically meaningful activations around objects and interactions of interest, while PyramidCLIP attends more to background regions.}
    \label{fig:gradcam}
\end{figure*}

\begin{figure*}[!t]
    \centering
    \includegraphics[width=\textwidth]{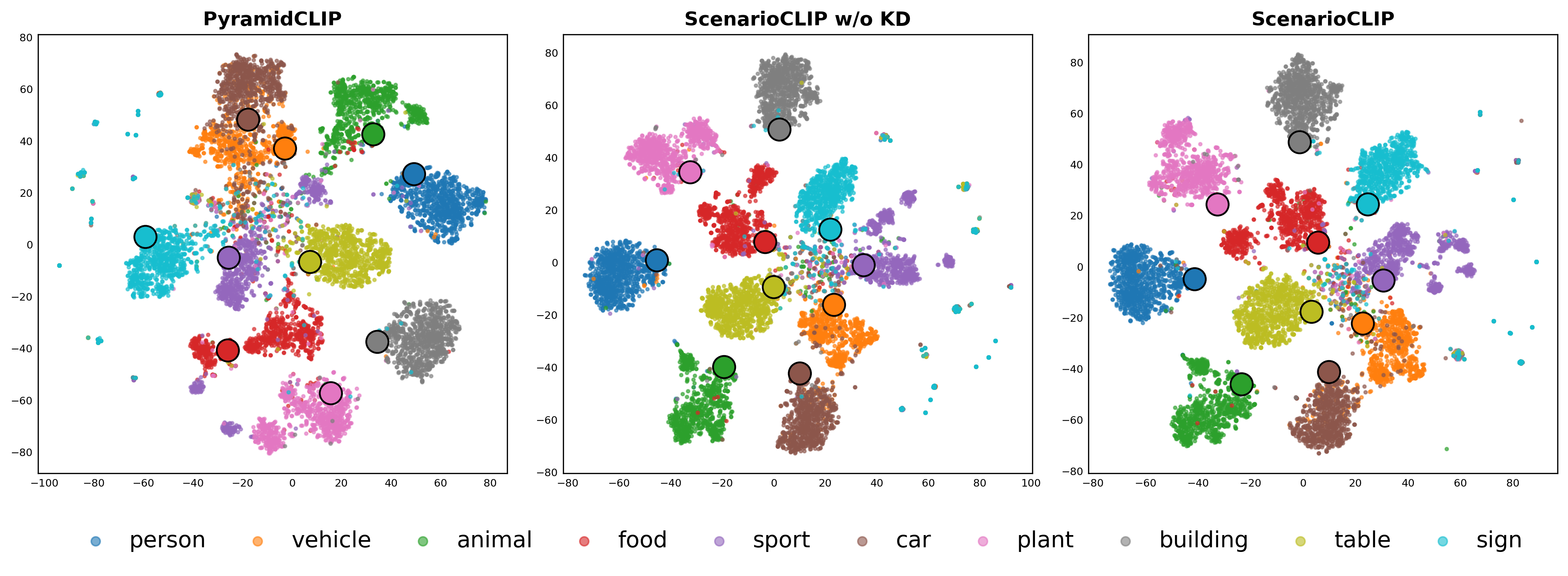}
    \caption{\textbf{t-SNE visualisation of object-level semantic 
    embeddings of 10{,}000 objects from the SCLARO dataset.} 
    Large points represent text features and small points represent 
    image features.}
    \label{fig:tsne}
\end{figure*}

\subsection{Relation Localisation}
\label{sec:rel_loc_details}

%To evaluate how well the relation encoder captures the spatial extent of an interaction, we train a lightweight decoder to predict a dense relation mask. For each annotated $(object_1, relation, object_2)$ triple, we construct a focused input image by zeroing out pixels outside the union of the $object_1$ and $object_2$ bounding boxes, and use the corresponding relation mask as supervision.

%The relation visual encoder serves as the visual backbone. We discard the class token, reshape the remaining patch tokens into an $h\times w$ feature map, and pass it through a small convolutional decoder (Conv$3\times3$-$256$-ReLU-Conv$3\times3$-$64$-ReLU-Conv$1\times1$-$1$) to produce patch-level logits, bilinearly upsampled to image resolution and sigmoid-activated. The decoder is optimised with an $\ell_2$ loss using Adam (learning rate $10^{-4}$). We evaluate two variants: \emph{Frozen} (encoder fixed, only decoder trained) and \emph{Trainable} (both updated).

%\Cref{tab:rel_localization} summarises relation localisation performance. With frozen encoders, PyramidCLIP~\cite{gao2022pyramidcliphierarchicalfeaturealignment} attains slightly higher Dice and IoU than ScenarioCLIP, consistent with its entangled backbone being trained jointly on global images, object crops, and relation masks. ScenarioCLIP achieves the lowest MAE. When encoders are updated, ScenarioCLIP w/o KD obtains the best Dice and IoU, while ScenarioCLIP with KD remains close behind, suggesting that the disentangled design does not compromise spatial localisation of interactions.

To benchmark how well a model's relation encoder captures the spatial
extent of an interaction grounded in the SCLARO dataset's focused-region
annotations, we train a lightweight decoder to predict a dense relation
mask. For each annotated $(object_1, relation, object_2)$ triple, we
construct a focused input image by zeroing out pixels outside the union
of the $object_1$ and $object_2$ bounding boxes, and use the
corresponding relation mask as supervision.
 
The relation visual encoder serves as the visual backbone. We discard
the class token, reshape the remaining patch tokens into an $h \times w$
feature map, and pass it through a small convolutional decoder
(Conv$3\times3$-256-ReLU-Conv$3\times3$-64-ReLU-Conv$1\times1$-1) to
produce patch-level logits, bilinearly upsampled to image resolution and
sigmoid-activated. The decoder is optimised with an $\ell_2$ loss using
Adam (learning rate $10^{-4}$). We evaluate two variants: \textit{Frozen}
(encoder fixed, only decoder trained) and \textit{Trainable} (both
updated).
 
\Cref{tab:rel_localization} summarises relation localisation
performance across all three models. With frozen encoders, PyramidCLIP
attains slightly higher Dice and IoU than ScenarioCLIP, while
ScenarioCLIP achieves the lowest MAE. When encoders are updated,
ScenarioCLIP w/o KD obtains the best Dice and IoU, while ScenarioCLIP
with KD remains close behind. Performance is broadly comparable across
all three models on this benchmark, indicating that the SCLARO
dataset's focused-region annotations support spatial localisation of
relations regardless of the specific encoder design used to learn from
them.

\begin{table*}[h]
\centering
\small
\begin{tabular}{l ccc ccc ccc}
\toprule
& \multicolumn{3}{c}{Actions} & \multicolumn{3}{c}{Objects} & \multicolumn{3}{c}{Relations} \\
\cmidrule(lr){2-4} \cmidrule(lr){5-7} \cmidrule(lr){8-10}
\textbf{KD Configuration} & Top-1 & Top-5 & Top-10 & Top-1 & Top-5 & Top-10 & Top-1 & Top-5 & Top-10 \\
\midrule
Unidirectional cross-level    & 57.13 & 74.66 & 79.82 & 49.45 & 63.93 & 68.06 & 19.30 & 35.34 & 43.70 \\
Bidirectional cross-level     & 56.70 & 74.21 & 79.39 & 49.14 & 63.61 & 67.76 & 18.92 & 35.44 & 43.75 \\
Symmetric                     & 56.89 & 74.47 & 79.75 & 46.99 & 63.24 & 67.52 & 19.24 & 35.42 & 43.81 \\
Asymmetric intra-modal (ours) & \textbf{57.63} & \textbf{75.02} & \textbf{79.99} & \textbf{51.86} & \textbf{64.43} & \textbf{68.30} & \textbf{19.56} & \textbf{37.03} & \textbf{45.86} \\
\bottomrule
\end{tabular}
\caption{Ablation of knowledge distillation configurations. Zero-shot image$\rightarrow$text retrieval Top-1/5/10 accuracy on the SCLARO dataset test set, across Actions, Objects, and Relations.}
\label{tab:ablation_kd}
\end{table*}

\begin{table*}[h]
\centering
\small
\begin{tabular}{l ccc ccc ccc}
\toprule
& \multicolumn{3}{c}{Actions} & \multicolumn{3}{c}{Objects} & \multicolumn{3}{c}{Relations} \\
\cmidrule(lr){2-4}\cmidrule(lr){5-7}\cmidrule(lr){8-10}
$\lambda_{KD}$ \textbf{schedule} & Top-1 & Top-5 & Top-10 & Top-1 & Top-5 & Top-10 & Top-1 & Top-5 & Top-10 \\
\midrule
Fixed-1            & \textbf{57.63} & 75.02 & 79.99 & 51.86 & 64.43 & 68.30 & \textbf{19.56} & \textbf{37.03} & 45.86 \\
Fixed-10           & 57.26 & 75.11 & 80.21 & 48.97 & 64.31 & 68.21 & 19.49 & 36.87 & 45.71 \\
Anneal $1\rightarrow 0$  & 57.15 & 75.23 & 80.39 & \textbf{52.28} & \textbf{64.52} & \textbf{68.40} & 19.47 & 37.03 & \textbf{45.95} \\
Anneal $10\rightarrow 0$ & 57.23 & \textbf{75.24} & \textbf{80.44} & 51.74 & 64.44 & 68.36 & 19.52 & 36.90 & 45.81 \\
\bottomrule
\end{tabular}
\caption{Zero-shot image$\rightarrow$text retrieval on the SCLARO dataset under different choices of the distillation weight $\lambda_{KD}$.}
\label{tab:lambda_ablation_zs}
\end{table*}

\subsection{Visualisations}
\label{sec:viz_details}

To understand what supervision from the SCLARO dataset's object
annotations encourages a model to attend to, we visualise patch-level
attention using Grad-CAM~\cite{Selvaraju_2019}. For each test
image, we feed the full frame through the object visual encoder and
interpret the cosine similarity between each patch token and the CLS
token as a proxy for patch relevance. These similarities are reshaped
into an $h \times w$ grid, upsampled to input resolution, and overlaid
on the original image using a jet colourmap.
 
\Cref{fig:gradcam} compares PyramidCLIP \cite{gao2022pyramidcliphierarchicalfeaturealignment}, ScenarioCLIP w/o KD, and
ScenarioCLIP across three images. ScenarioCLIP produces sharper, more
localised responses around semantically meaningful regions: the batter,
bat, and catcher in the first image, the camera and child's hands in the
second, and both people and their interaction in the third. PyramidCLIP
attends more to background regions in each case.
 
To assess the learned representation space qualitatively, we visualise
object-level semantic embeddings of $10{,}000$ objects from the SCLARO
dataset test and validation sets using
t-SNE~\cite{JMLR:v9:vandermaaten08a} (\cref{fig:tsne}). PyramidCLIP's
clusters are overlapping and loosely formed. ScenarioCLIP w/o KD shows
clearer separation but relatively sparse clusters. ScenarioCLIP produces
compact, well-organised clusters with minimal overlap, with semantically
similar categories such as (car, vehicle) positioned in close proximity.
This pattern is consistent with the asymmetric intra-modal distillation
acting as a geometric regulariser on the representation learned from the
dataset's object-level annotations.

%\begin{figure*}[!t]
%  \centering
%  \begin{subfigure}[b]{0.32\textwidth}
%    \includegraphics[width=\linewidth]{fig/tsne_pc.png}
 %   \caption{PyramidCLIP}\label{fig:tsne:a}
%  \end{subfigure}\hfill
%  \begin{subfigure}[b]{0.32\textwidth}
%    \includegraphics[width=\linewidth]{fig/tsne_sc0.png}
%    \caption{ScenarioCLIP w/o KD}\label{fig:tsne:b}
%  \end{subfigure}\hfill
 % \begin{subfigure}[b]{0.32\textwidth}
 %   \includegraphics[width=\linewidth]{fig/tsne_sc1.png}
%    \caption{ScenarioCLIP}\label{fig:tsne:c}
%  \end{subfigure}
 % \caption{\textbf{t-SNE visualisation of object-level semantic 
%  embeddings of 10{,}000 objects from the SCLARO Dataset.} 
%  Large points represent text features and small points represent 
 % image features.}
%  \label{fig:tsne}
%\end{figure*}

\subsection{KD Ablations: Direction and Weight}
\label{sec:kd_ablations}
 
As described in the main paper, the total objective is:
\begin{equation}
    \mathcal{L}_{\text{total}} = \lambda_{KD} \cdot \mathcal{L}_{KD} + \lambda_{CA} \cdot \mathcal{L}_{CA}.
\end{equation}
We study two orthogonal design choices in this objective: the
\textit{direction} of the distillation term $\mathcal{L}_{KD}$ (which
embeddings teach which), and the \textit{weighting schedule} of
$\lambda_{KD}$ over training. All main paper results use the asymmetric
intra-modal direction with $\lambda_{KD}=1$ fixed throughout (Fixed-1).
 
\paragraph{Direction.} We further investigate two alternative KD
configurations: \textit{cross-level KD}, which adds distillation between
global and fine-grained embeddings across modalities, and
\textit{symmetric KD}, which applies distillation in both directions at
every level. For cross-level KD, we evaluate both a \textit{bidirectional}
variant, where the global visual teacher distills into fine-grained text
students and fine-grained text teachers distill back into the global
visual student, and a \textit{unidirectional} variant, where only the
global-visual-to-text direction is retained. All three alternative
configurations underperform our asymmetric intra-modal design
(\cref{tab:ablation_kd}).

Cross-level KD introduces competing gradient signals on $V_G$ from both
the peer-level contrastive and cross-level distillation objectives. This
conflict is more pronounced in the bidirectional variant, where the
reverse direction (fine-grained text teaching $V_G$) directly competes
with the existing $\mathrm{Contrastive}(V_G, T_G)$ objective at the
global level. Removing this reverse direction in the unidirectional
variant recovers part of the gap, though it still trails our asymmetric
intra-modal design. Symmetric KD similarly conflicts with the peer-level
contrastive objective at the global visual level. These results
empirically validate the asymmetric design choice described in the main
paper.
 
\paragraph{Weight.} We study the effect of $\lambda_{KD}$ under four
settings: \textbf{Fixed-1} ($\lambda_{KD}=1$ throughout), \textbf{Fixed-10}
($\lambda_{KD}=10$ throughout), \textbf{Anneal $1\rightarrow 0$}, and
\textbf{Anneal $10\rightarrow 0$}. For the annealed variants, training
progress is parameterised as $p=(e+1)/E \in [0,1]$ with breakpoints
$p_1=0.4$ and $p_2=0.7$:
\begin{equation}
\resizebox{\linewidth}{!}{$
\lambda_{KD}(p) =
\begin{cases}
\lambda_{\text{start}}, & p \le p_1,\\[4pt]
\lambda_{\text{mid}} + \frac{\lambda_{\text{start}}-\lambda_{\text{mid}}}{2}\bigl(1+\cos(\pi t_1)\bigr), & p_1 < p \le p_2,\\[4pt]
\lambda_{\text{end}} + \frac{\lambda_{\text{mid}}-\lambda_{\text{end}}}{2}\bigl(1+\cos(\pi t_2)\bigr), & p > p_2,
\end{cases}
$}
\label{eq:lambda-sched}
\end{equation}
where $t_1=\tfrac{p-p_1}{p_2-p_1}$ and $t_2=\tfrac{p-p_2}{1-p_2}$, with
$\lambda_{\text{end}}=0$, $(\lambda_{\text{start}},\lambda_{\text{mid}})=(1,0.5)$
for Anneal $1\rightarrow 0$ and $(10,1)$ for Anneal $10\rightarrow 0$.
Zero-shot retrieval results for all four settings are shown in
\cref{tab:lambda_ablation_zs}.